\documentclass[journal]{IEEEtran}

\usepackage{amsmath,amssymb}
\usepackage{booktabs}
\usepackage{graphicx}
\usepackage{subcaption}
\usepackage[table]{xcolor}
\usepackage{multirow}
\usepackage{cite}
\usepackage{hyperref}
\usepackage{enumitem}

\begin{document}

\title{WATCH: Wide-Area Archaeological Site Tracking for Change Detection}

\author{Girmaw~Abebe~Tadesse$^{*}$,~\IEEEmembership{}
        Titien~Bartette,
        Andrew~Hassanali,
        Allen~Kim,
        Jonathan~Chemla,
        Andrew~Zolli,
        Yves~Ubelmann,
        Caleb~Robinson,
        Inbal~Becker-Reshef,
        and~Juan~Lavista~Ferres%
\thanks{G.~A.~Tadesse, A.~Kim, C.~Robinson, I.~Becker-Reshef, and J.~Lavista~Ferres are with the Microsoft AI for Good Research Lab.}%
\thanks{T.~Bartette, J.~Chemla, and Y.~Ubelmann are with Iconem, Paris, France.}%
\thanks{A.~Hassanali and A.~Zolli are with Planet Labs PBC.}%
\thanks{$^{*}$Corresponding author: Girmaw Abebe Tadesse (e-mail: gtadesse@microsoft.com).}%
}

\maketitle

\begin{abstract}
Monitoring archaeological sites at scale is vital for protecting cultural heritage, yet pinpointing when disturbances occur remains difficult because visual cues are subtle and ground-truth data are sparse.
We introduce WATCH, a framework for month-level change-event localization over PlanetScope satellite mosaics (2017--2024, $4.7\,\mathrm{m/px}$) that supports three complementary scoring approaches: (i)~Temporal Embedding Distance (TED), a training-free method that scores month-to-month deviations from a local temporal reference; (ii)~Self-Supervised Change Detection (SSCD), an ensemble of reconstruction, forecasting, and latent-novelty signals; and (iii)~a Weakly Supervised (WS) temporal localization model trained with sparse event-month labels.
We benchmark WATCH on 1,943 archaeological sites in Afghanistan using embeddings from six foundation models (CLIP, GeoRSCLIP, SatMAE, Prithvi-EO-2.0, DINOv3, and Satlas-Pretrain) alongside a handcrafted spectral and texture baseline, and assess cross-regional generalization on sites in Syria, Turkey, Pakistan, and Egypt.
The unsupervised approaches (TED, SSCD) consistently outperform the weakly supervised alternative. TED with SatMAE achieves the highest exact-month recall (55\% at $m{=}0$), while TED with GeoRSCLIP, CLIP, or Satlas-Pretrain reaches 92.5\% within a three-month tolerance ($m{=}3$). Handcrafted features remain competitive for exact-month detection under weak supervision.
Our directional margin analysis reveals systematic temporal biases: SSCD paired with GeoRSCLIP or Prithvi-EO-2.0 exhibits the strongest early-warning profile, detecting anomalies before the recorded event, while TED favors confirmation-oriented detection after a change has materialized.
These results show that satellite imagery combined with foundation-model embeddings enables scalable, decision-relevant heritage monitoring. Code: \url{https://github.com/microsoft/WATCH}.
\end{abstract}


\section{Introduction}
\IEEEPARstart{A}{rchaeological} sites are irreplaceable records of human history and cultural identity, yet face growing threats from looting, unregulated construction, agricultural expansion, and conflict-driven destruction.
Advances in earth observation~(EO) now provide routine satellite imagery at monthly or sub-monthly intervals, establishing remote sensing as a key tool for heritage monitoring~\cite{parcak2016satellite-evidence, cuca2023monitoring, negula2015earth, levin2019world}.
Because many sites are remote or situated in unstable environments, satellite-based monitoring has become indispensable for large-scale cultural heritage preservation~\cite{tapete2019looting,agapiou2021unesco}.

For operational decisions, heritage practitioners require temporal localization of change events, not merely binary site-level classification, since accurate detection is critical for timely intervention. Satellite-based looting detection has a long history, but early approaches relied on visual inspection, sometimes augmented by simple image processing~\cite{menze2012mapping,tapete2019looting}.
These manual workflows are labor-intensive, subjective, and difficult to scale.
Looting leaves subtle, spatially dispersed traces, including disturbed soil and localized spectral anomalies, that are easily confounded with natural processes such as erosion or agriculture.
Machine learning offers a scalable alternative~\cite{vincent2025detecting,tadesse2026satellite}, yet automated systems still struggle to pinpoint \emph{when} disturbances occur for three main reasons: ground-truth labels are sparse and often temporally imprecise; changes are visually subtle and heterogeneous, easily confounded by seasonal effects, illumination variation, and sensor noise; and large-scale monitoring must operate efficiently across thousands of sites with limited computational and annotation budgets.
We address these challenges through \emph{site-centric} monitoring, representing each site as a monthly time series of feature embeddings extracted from a masked (if available) spatial grid.

Prior work on archaeological site monitoring has focused on binary classification, distinguishing looted from preserved sites using single images or full time series~\cite{vincent2025detecting,tadesse2026satellite}.
Recent geospatial foundation models, including SatMAE~\cite{cong2022satmae}, SatCLIP~\cite{klemmer2024satclip}, DINOv3~\cite{simeoni2025dinov3}, and Prithvi-EO~\cite{jakubik2023foundation}, have advanced performance on a range of downstream EO tasks, yet their effectiveness for temporal change detection at archaeological sites remains largely unexplored. We benchmark six foundation models in this study.
Handcrafted features (e.g., spectral indices and texture descriptors) can capture looting-related patterns, but have not been rigorously compared to foundation-model embeddings in this setting.
More broadly, change detection in remote sensing has been studied extensively, typically under fully supervised or bi-temporal protocols requiring dense annotations or paired imagery~\cite{daudt2018fully,chalapathy2019deep}.
These methods are not directly applicable to our setting, which requires monthly temporal resolution under weak or absent supervision over long time series.
\emph{Month-level change detection} for archaeological sites, however, remains underexplored.
No existing change detection framework for archaeological sites jointly (i)~operates on monthly satellite time series, (ii)~exploits modern foundation-model embeddings, (iii)~functions under minimal or noisy supervision, and (iv)~evaluates temporal tolerance in a manner relevant to real-world heritage monitoring.

\begin{figure*}[!t]
    \centering
    \includegraphics[width=\textwidth]{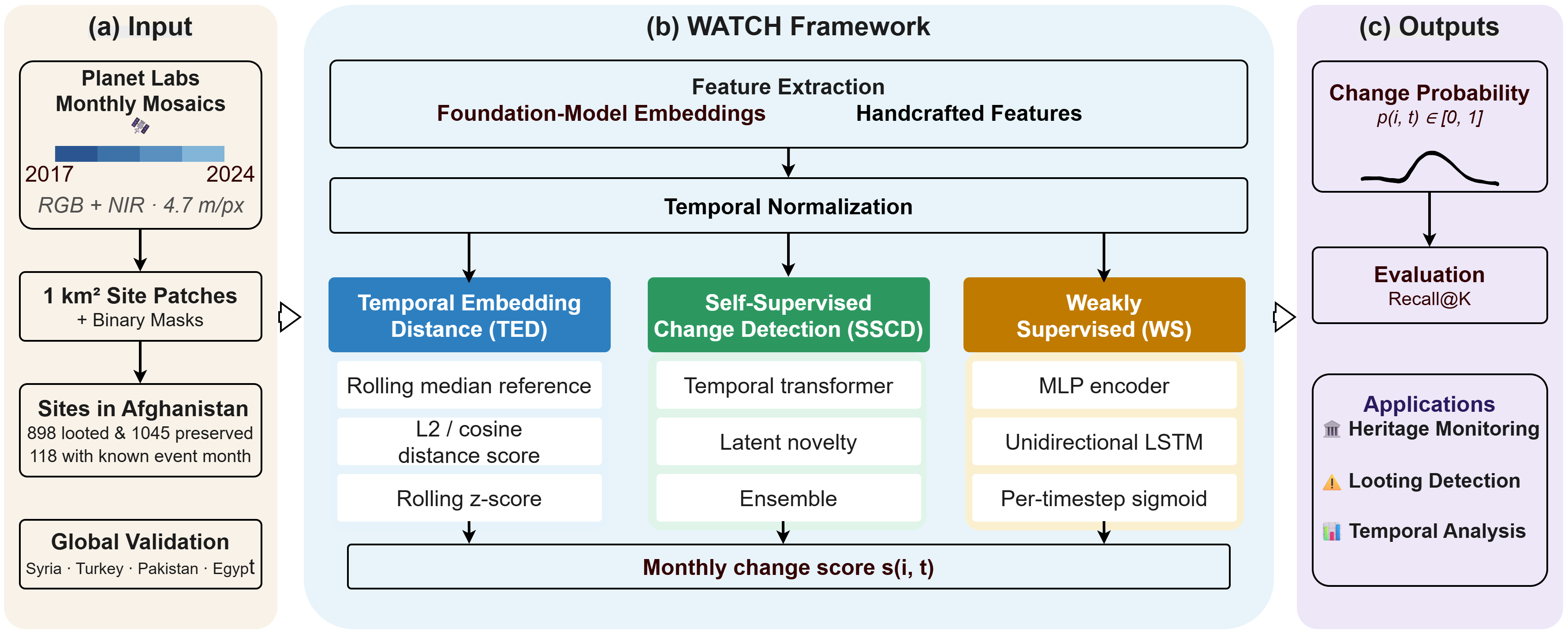}
    \caption{Overview of the WATCH framework. (a)~PlanetScope monthly mosaics (RGB+NIR, 4.7\,m/px, 2017--2024) are processed into 1\,km\textsuperscript{2} site patches with spatial masks for 1,943 archaeological sites in Afghanistan, and later validated across sites beyond Afghanistan, such as Syria, Turkey, Pakistan, and Egypt. (b)~Foundation-model embeddings and handcrafted features are extracted and temporally normalized, then scored via three complementary approaches: TED (Temporal Embedding Distance), SSCD (Self-Supervised Change Detection), and WS (Weakly Supervised), each producing a monthly change score. (c)~Scores are converted to change probabilities and evaluated using recall with symmetric and directional temporal margins, supporting heritage monitoring, looting detection, and temporal analysis applications.}
    \label{fig:watch_framework}
\end{figure*}

We present \textbf{WATCH} (\textbf{W}ide-area \textbf{A}rchaeological site \textbf{T}racking for \textbf{CH}ange detection), a framework for month-level change detection in archaeological site monitoring using PlanetScope monthly mosaics (2017--2024, $4.7\,\mathrm{m/px}$) (see Fig.~\ref{fig:watch_framework}).
WATCH operates on per-site time series of geospatial foundation-model embeddings and supports three complementary scoring approaches: (i)~Temporal Embedding Distance (TED), a training-free deviation score; (ii)~Self-Supervised Change Detection (SSCD), an ensemble of reconstruction, forecasting, and novelty signals; and (iii)~a Weakly Supervised (WS) temporal localization model trained with sparse, truncated event-month labels.
We benchmark embeddings from Satlas-Pretrain~\cite{bastani2023satlaspretrain}, GeoRSCLIP~\cite{zhang2024georsclip}, SatMAE~\cite{cong2022satmae}, CLIP~\cite{radford2021learning}, DINOv3~\cite{simeoni2025dinov3}, and Prithvi-EO-2.0~\cite{szwarcman2025prithvi} against handcrafted features, and introduce a unified evaluation protocol based on Recall@$K$ (hereafter simply \emph{recall}) with symmetric and directional temporal margins with the top-K months based on their likelihood of change.
We validate on sites in Afghanistan and assess cross-regional transfer on sites in Syria, Turkey, Pakistan, and Egypt (see Fig.~\ref{suppfig:global_and_countries}).

Our contributions are:
\begin{enumerate}[leftmargin=*]
    \item Formulating archaeological site monitoring as monthly change-event detection over site-centric satellite time series.
    \item Proposing WATCH with three complementary scoring approaches (TED, SSCD, and WS) spanning unsupervised and weakly supervised regimes.
    \item Systematically benchmarking six foundation-model embeddings against handcrafted features for this task.
    \item Introducing a temporally tolerant evaluation protocol with directional margin analysis.
\end{enumerate}
To promote reproducibility and support continued research, the codebase has been made publicly accessible at \url{https://github.com/microsoft/WATCH}.

\section{Dataset}
\label{sec:dataset}

\begin{figure}[!t]
    \centering
    \begin{subfigure}[t]{\columnwidth}
        \centering
        \includegraphics[width=\columnwidth]{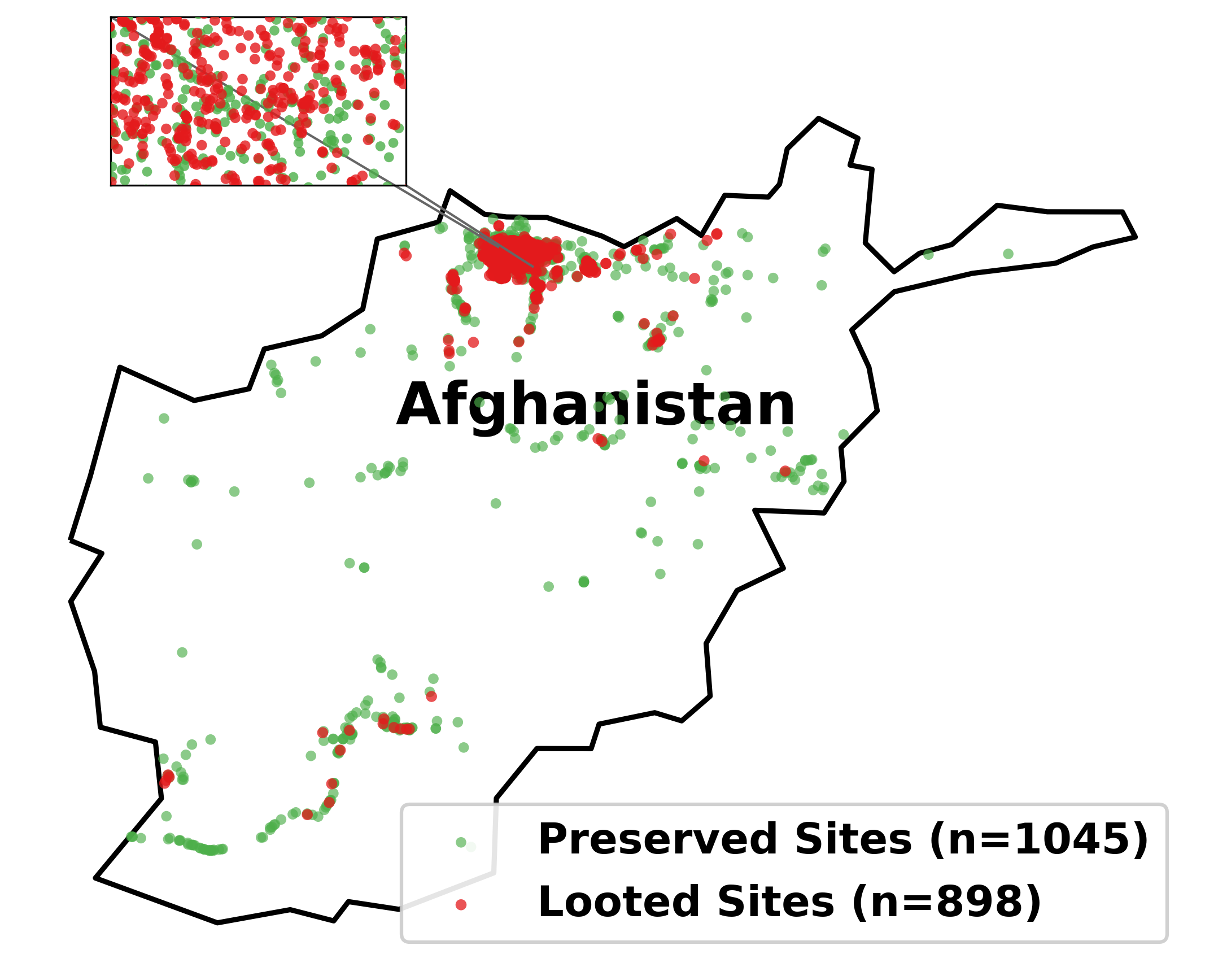}
        \caption{}
        \label{fig:sites}
    \end{subfigure}

    \begin{subfigure}[t]{\columnwidth}
        \centering
        \includegraphics[width=\columnwidth]{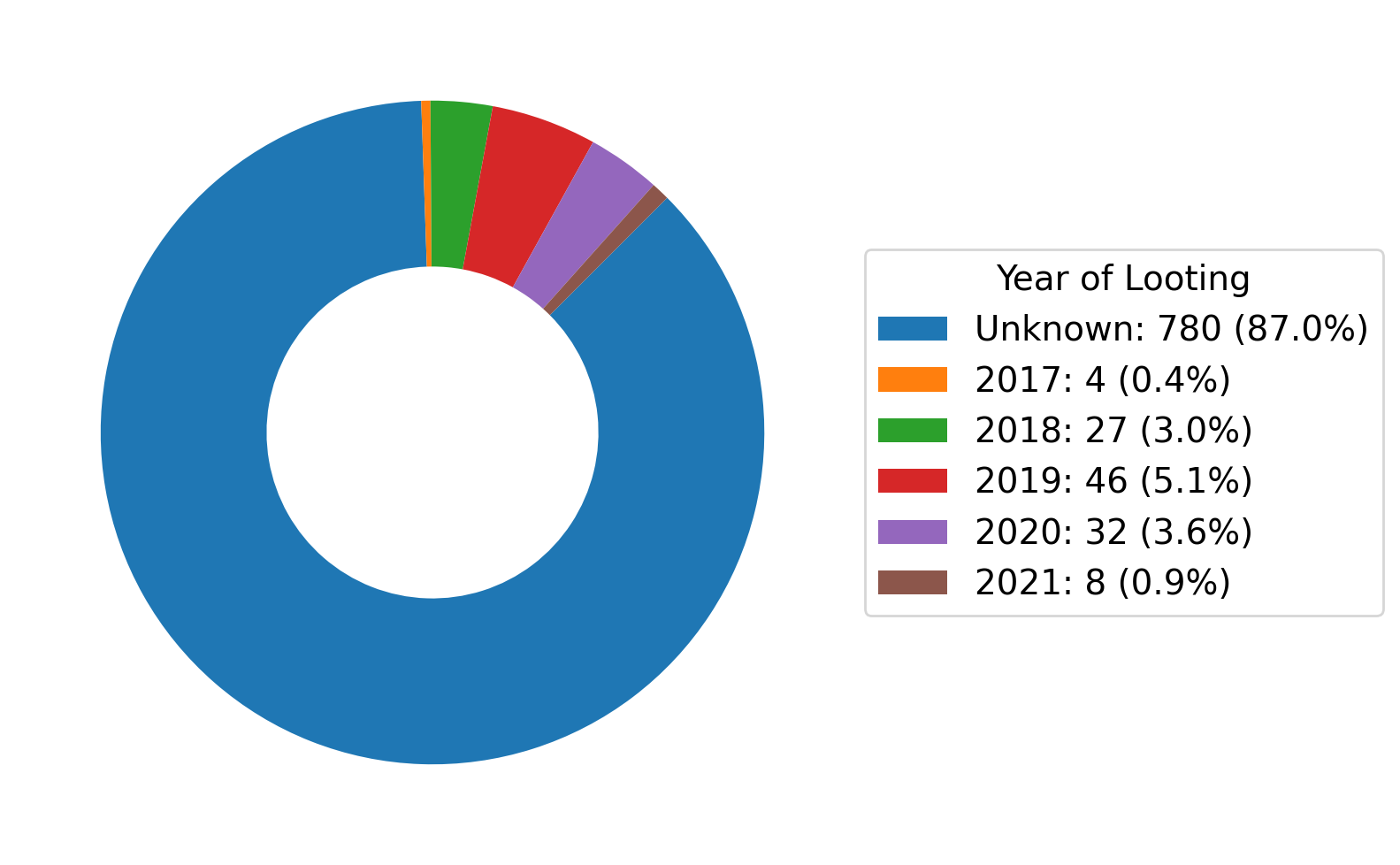}
        \caption{}
        \label{fig:looted_month}
    \end{subfigure}

    \begin{subfigure}[t]{\columnwidth}
    \centering
        \includegraphics[width=0.49\columnwidth]{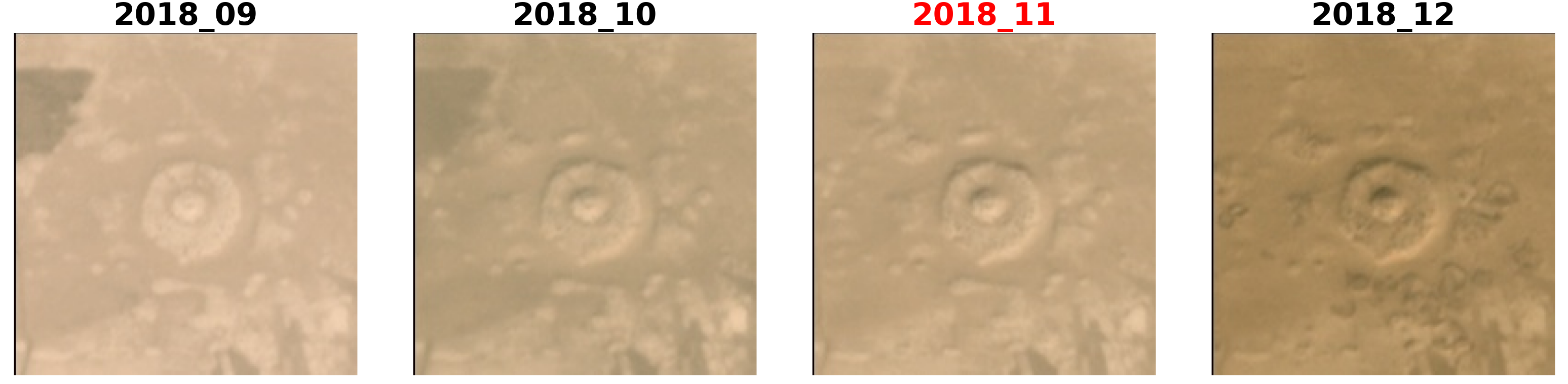}
        \includegraphics[width=0.49\columnwidth]{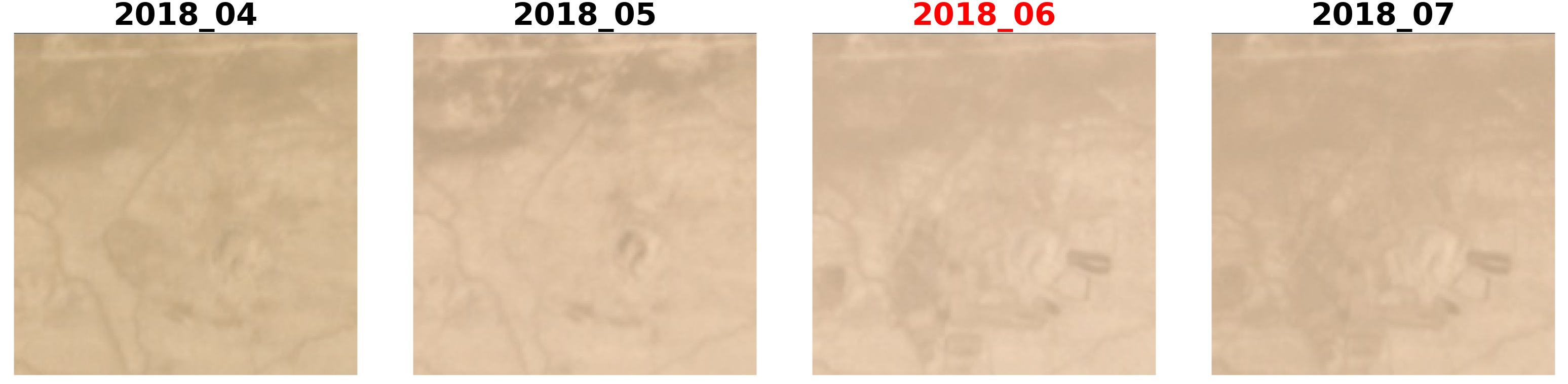}
        \includegraphics[width=0.49\columnwidth]{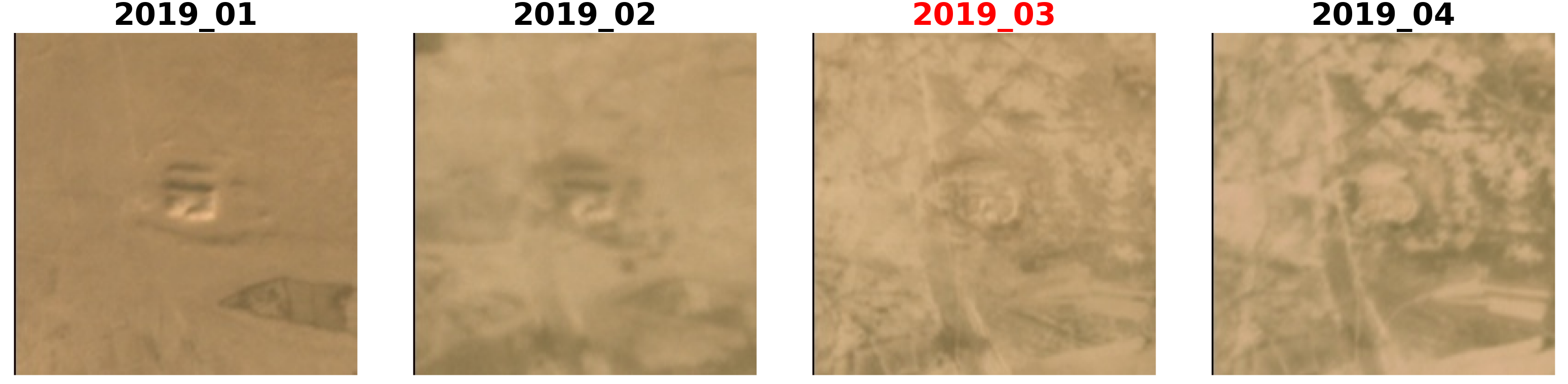}
        \includegraphics[width=0.49\columnwidth]{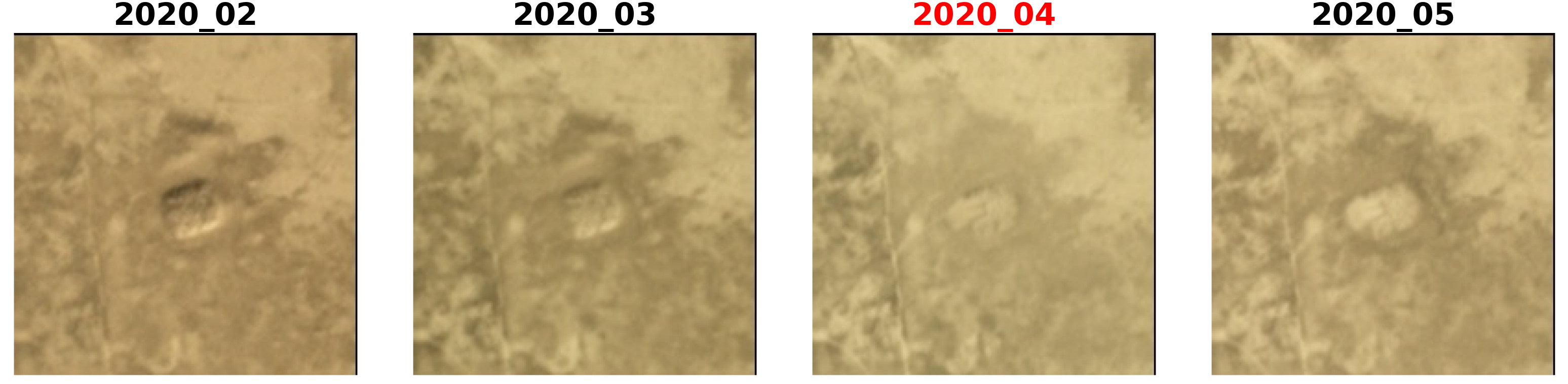}
        \caption{}\label{fig:looted_month_examples}
    \end{subfigure}

    \caption{(a)~Overview of the archaeological sites in Afghanistan used in this study, comprising $1{,}045$ \textcolor{green}{preserved} and $898$ \textcolor{red}{looted} sites.
    (b)~Stratification of looted sites by the month in which looting was first detected. Only 117 of 898~looted sites have a reliably observed month of looting activity. The majority of documented events occurred prior to 2021. (c)~Examples of Afghanistan archaeological sites for which the month of looting is known. The mosaics from the two months preceding the observed looting, \textcolor{red}{the month of reported looting}, and the post-looting month are shown to illustrate the temporal progression of looting activity.}
    \label{fig:dataset_overview}
\end{figure}

Our primary dataset comprises 1,943 archaeological sites in Afghanistan: 898 looted (46.2\%) and 1,045 preserved (53.8\%) as shown in Fig.~\ref{fig:sites}.
Sites were compiled from the \textit{Archaeological Gazetteer of Afghanistan}~\cite{ball2019archaeological} and subsequently verified and expanded by expert archaeologists using high-resolution satellite imagery platforms (Google Earth, ESRI, Bing) to classify each site as ``looted'' or ``preserved'' based on evidence of deliberate damage before 2023.
Of the looted sites, only 117 ($\sim$13\%) have a reliably known month of looting activity (see Fig.~\ref{fig:looted_month}). Examples of sites before and after a reported month of looting are shown in Fig.~\ref{fig:looted_month_examples}.
For cross-regional validation, we include additional sites from Syria, Turkey, Pakistan, and Egypt (see Figs.~\ref{suppfig:global_and_countries} and~\ref{suppfig:global_countries_detail} in the Supplementary Material).

For each site and month, we use PlanetScope monthly mosaics (RGB+NIR) at 4.7~m/pixel spanning January 2017 to December 2024 (96 months), constructing median composite patches of approximately 1~km$\times$1~km centered on each site (see Table~\ref{tab:qual_examples}).
We manually annotate polygonal site footprints and rasterize them into binary spatial masks aligned to each patch to suppress non-site context (roads, settlements, agricultural fields).

\begin{table*}[!t]
    \centering
    \caption{PlanetScope monthly mosaics and corresponding binary masks for selected sites in Afghanistan, shown for December of each year from 2017 to 2024. Site coordinates were provided by collaborators at \href{https://iconem.com/}{Iconem}. For each site, a 1~km$\times$1~km monthly mosaic was retrieved. Masks were created through manual annotation. Looted sites are characterized by disturbed soil tones and textures, whereas preserved sites display relatively uniform surface patterns.}
    \label{tab:qual_examples}
    \resizebox{\textwidth}{!}{
    \begin{tabular}{c c c c c c c c c c}
        & \textbf{2017} & \textbf{2018} & \textbf{2019} & \textbf{2020} & \textbf{2021} & \textbf{2022} & \textbf{2023} & \textbf{2024} &\textbf{Mask} \\
        \multirow{1}{*}{Preserved}&
        \includegraphics[width=0.09\textwidth]{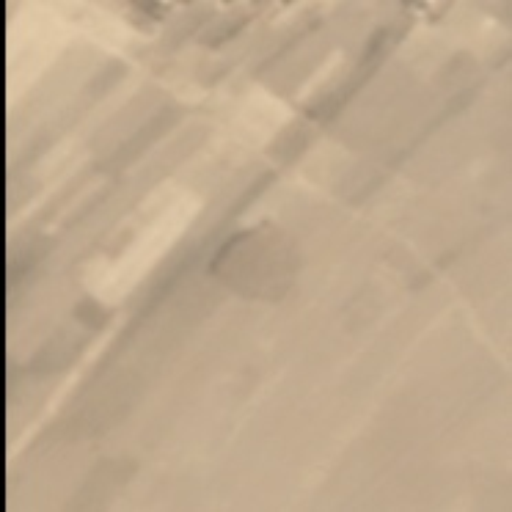} &
        \includegraphics[width=0.09\textwidth]{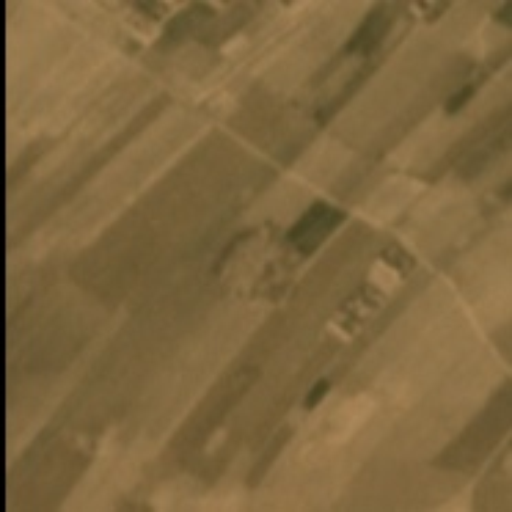} &
        \includegraphics[width=0.09\textwidth]{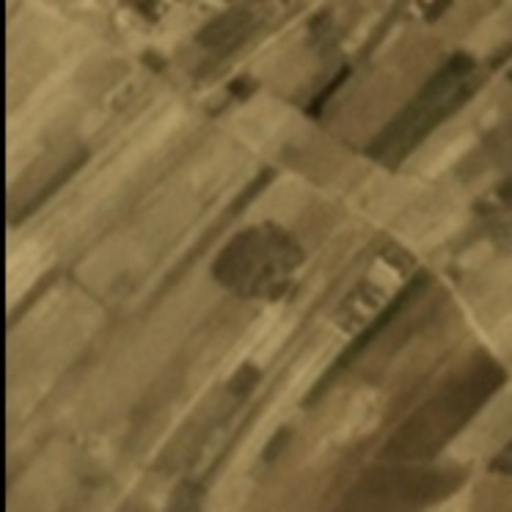} &
        \includegraphics[width=0.09\textwidth]{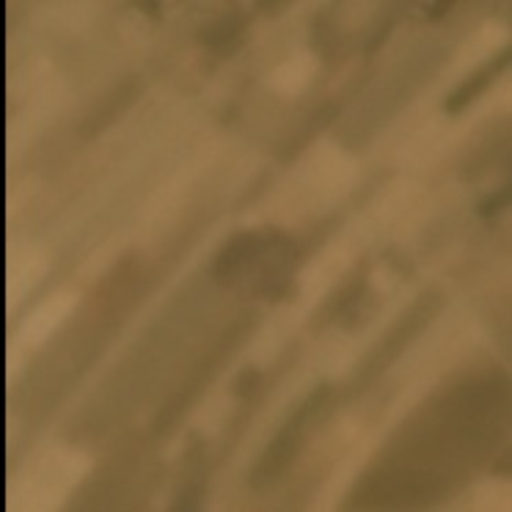} &
        \includegraphics[width=0.09\textwidth]{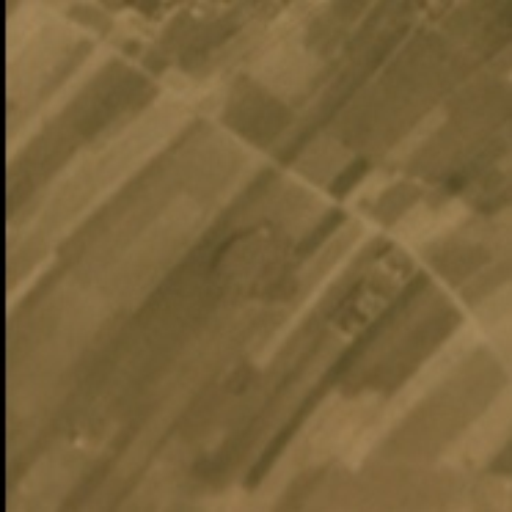} &
        \includegraphics[width=0.09\textwidth]{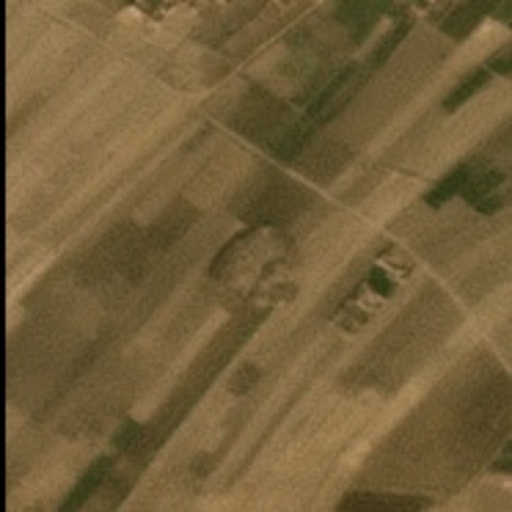} &
        \includegraphics[width=0.09\textwidth]{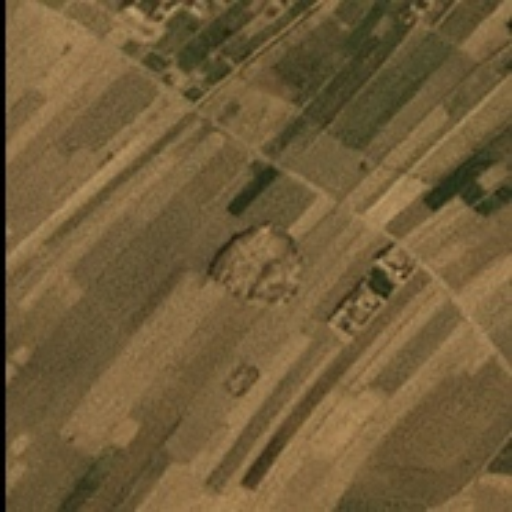} &
        \includegraphics[width=0.09\textwidth]{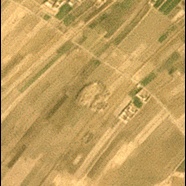} &
        \includegraphics[width=0.09\textwidth]{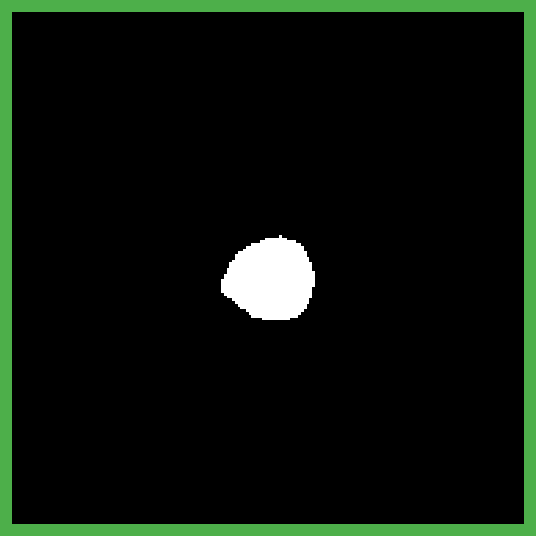} \\ \hline
           \multirow{1}{*}{Looted} &
        \includegraphics[width=0.09\textwidth]{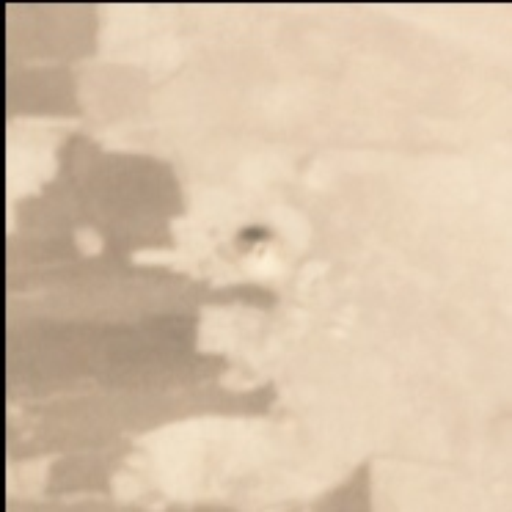} &
        \includegraphics[width=0.09\textwidth]{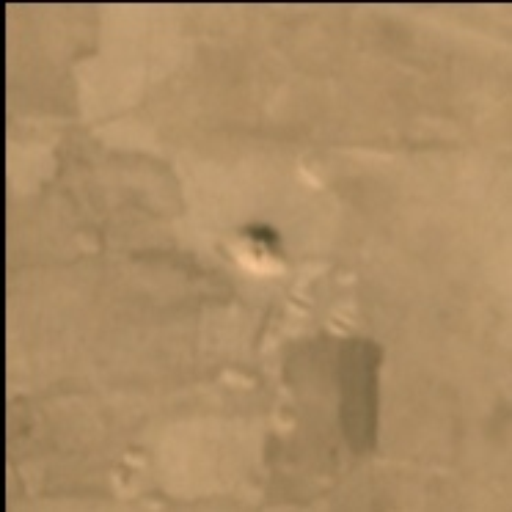} &
        \includegraphics[width=0.09\textwidth]{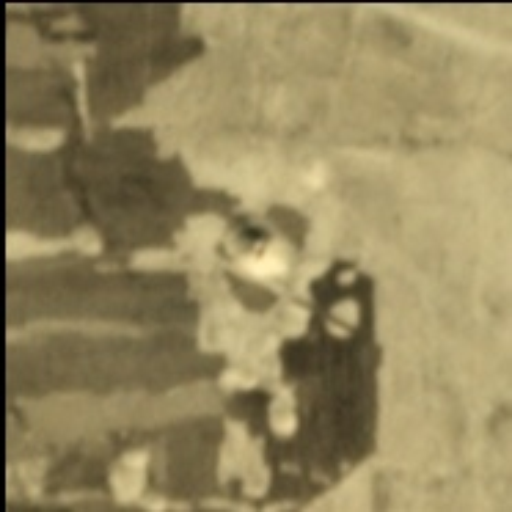} &
        \includegraphics[width=0.09\textwidth]{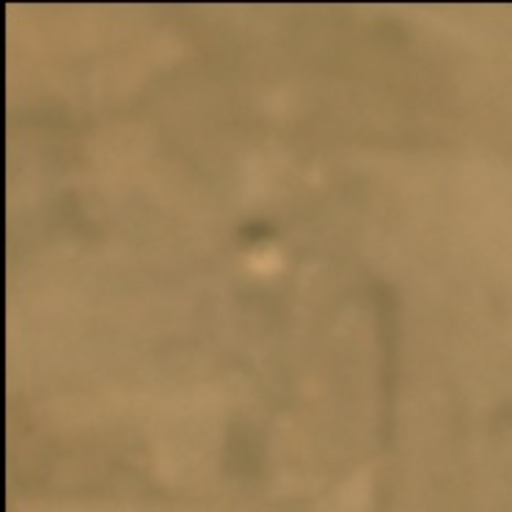} &
        \includegraphics[width=0.09\textwidth]{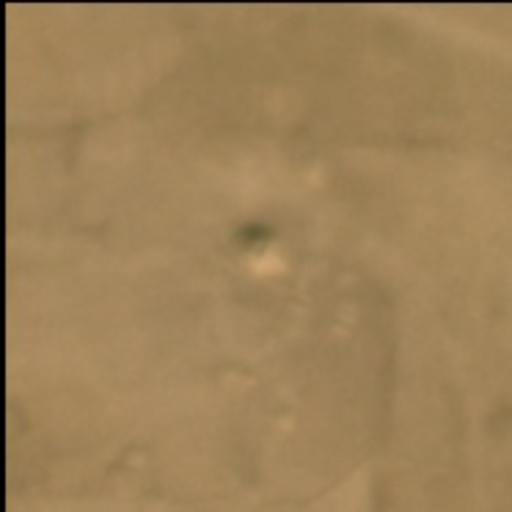} &
        \includegraphics[width=0.09\textwidth]{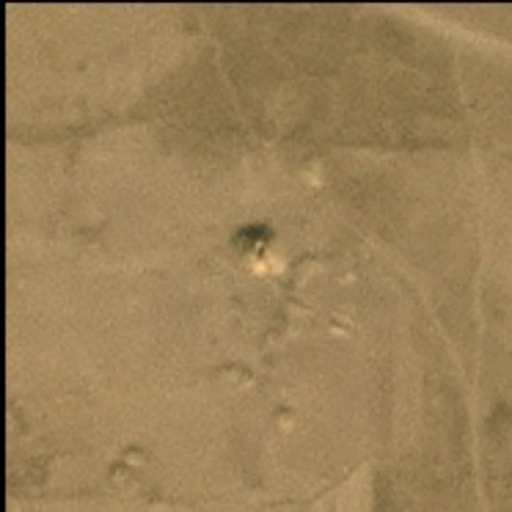} &
        \includegraphics[width=0.09\textwidth]{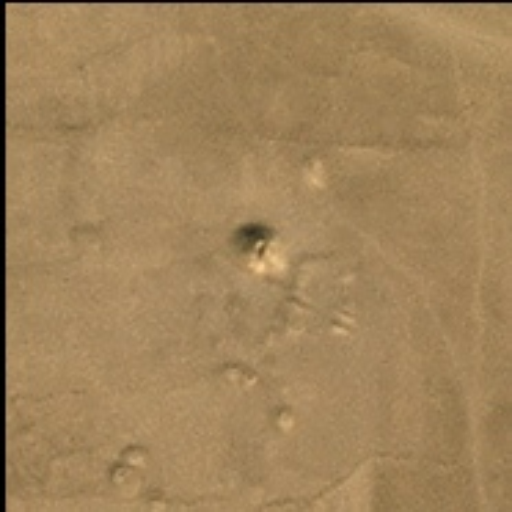} &
        \includegraphics[width=0.09\textwidth]{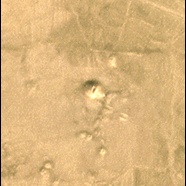} &
        \includegraphics[width=0.09\textwidth]{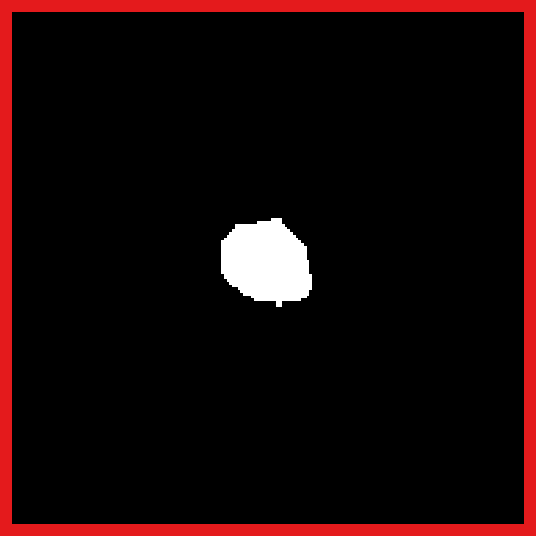}
    \end{tabular}
    }
\end{table*}

\section{Methodology}
\label{sec:methodology}

\subsection{Problem Formulation}
We monitor each archaeological site over $T=96$ months (January 2017 to December 2024).
For each site $i$ and month $t$, we extract an embedding $\mathbf{z}_{i,t}\in\mathbb{R}^{d}$ from the corresponding PlanetScope mosaic using a frozen foundation model or handcrafted feature pipeline (Section~\ref{sec:feature_extraction}).
The objective is to produce a probability $p_{i,t}\in[0,1]$ indicating the likelihood that a change event (e.g., looting onset) occurred at month~$t$.

\subsection{Feature Extraction}
\label{sec:feature_extraction}

For each site and month we extract a feature vector from the masked $1\,\text{km}^2$ PlanetScope mosaic using one of seven embedding/feature families; a detailed account of the extraction pipeline is given in~\cite{tadesse2026satellite}.
We select six foundation-model embeddings spanning vision--language alignment, self-supervised pretraining, and supervised large-scale pre-training, alongside a handcrafted feature baseline.
\textbf{CLIP}~\cite{radford2021learning} uses the frozen ViT-B/16 image encoder trained with contrastive language--image pre-training on natural images, producing a 512-dimensional embedding.
\textbf{GeoRSCLIP}~\cite{zhang2024georsclip} adapts the CLIP framework to remote sensing via the RS5M dataset and an OpenCLIP ViT-B/32 backbone, yielding 512-dimensional features.
\textbf{Prithvi-EO-2.0}~\cite{szwarcman2025prithvi} is a 600\,M-parameter geospatial foundation model based on a custom Vision Transformer that natively ingests four-band (BGRN) input with temporal position encoding; the CLS token is projected to a 1024-dimensional embedding.
\textbf{SatMAE}~\cite{cong2022satmae} pre-trains a ViT-Base encoder with masked autoencoders on multi-spectral satellite imagery using a group-channel masking strategy, producing 768-dimensional features.
\textbf{Satlas-Pretrain}~\cite{bastani2023satlaspretrain} employs a ResNet-152 backbone pre-trained on large-scale satellite imagery; we extract 2048-dimensional features from the globally average-pooled final convolutional layer.
\textbf{DINOv3}~\cite{simeoni2025dinov3} applies self-supervised learning via a ViT-Large/16 encoder trained on the SAT-493M satellite dataset, producing 1024-dimensional embeddings.
Finally, we include a \textbf{handcrafted} baseline comprising 60 statistical, spectral-index, and textural features (e.g., NDVI, NDWI, GLCM texture, LBP) computed directly from the four-band imagery without any learned model.
The resulting feature dimensions range from 60 (Handcrafted) to 2048 (Satlas-Pretrain), and all embeddings are extracted from a frozen encoder with no task-specific fine-tuning.

\subsection{Feature Normalization and Missingness}
\label{sec:normalization}

To mitigate scale mismatch across embeddings and seasonal variation, we apply a two-stage normalization.
The first stage removes seasonal effects: for each embedding dimension, we standardize by the mean and standard deviation of that dimension computed over all sites and years \emph{for the same calendar month}, yielding
\begin{equation}
\tilde{\mathbf{z}}_{i,t}
= \frac{\mathbf{z}_{i,t} - \boldsymbol{\mu}^{(m(t))}}{\boldsymbol{\sigma}^{(m(t))} + \varepsilon},
\end{equation}
where $m(t) \in \{1,\dots,12\}$ is the calendar month of $t$, $\boldsymbol{\mu}^{(m)}, \boldsymbol{\sigma}^{(m)} \in \mathbb{R}^d$ are the month-specific statistics, and $\varepsilon>0$ ensures numerical stability.
Because this standardization operates independently within each of the 12 calendar months, the pooled distribution of $\tilde{\mathbf{z}}_{i,t}$ across all 96 months may still have non-zero mean and non-unit variance due to residual differences between calendar months and unequal sample counts.
A second, global standardization step corrects for this:
\begin{equation}
\mathbf{z}'_{i,t}
= \frac{\tilde{\mathbf{z}}_{i,t} - \boldsymbol{\mu}}{\boldsymbol{\sigma} + \varepsilon},
\end{equation}
where $\boldsymbol{\mu}, \boldsymbol{\sigma} \in \mathbb{R}^d$ are the mean and standard deviation over all $\tilde{\mathbf{z}}_{i,t}$ regardless of calendar month.
Fewer than 0.5\% of site--month pairs have missing imagery (primarily due to acquisition gaps); these are imputed with the corresponding calendar-month mean prior to normalization.

\subsection{Unsupervised and Weakly Supervised Change Scoring}
\label{sec:scoring}

We employ three complementary scoring approaches trading supervision for flexibility. The two unsupervised approaches are label-agnostic and designed for stability across embeddings and deployment settings: \textbf{TED} (Temporal Embedding Distance) scores month-to-month deviations from a local temporal reference, and \textbf{SSCD} (Self-Supervised Change Detection) combines reconstruction, forecasting, and latent-novelty signals into an ensemble score. We use ``unsupervised'' to refer to these two approaches because they require no looting-event labels for scoring, acknowledging that the underlying embeddings were pretrained with various forms of supervision or self-supervision. The third approach, a \textbf{Weakly Supervised (WS)} temporal localization model, can yield sharper temporal localization when sufficient known-month labels exist, but requires training and careful handling of label sparsity.

\subsubsection{Temporal Embedding Distance (TED)}
TED scores the deviation of month~$t$ from a robust reference computed over recent history.
For target month $t$, let $\mathcal{P}(t)$ denote the set of up to $R=3$ preceding months; when fewer than $R$ months precede~$t$ (i.e., the first months of the series), the median is computed over the available subset.
The reference vector is the coordinate-wise median of available normalized embeddings:
\begin{equation}
\mathbf{B}_{i,t}=\operatorname{median}\left(\{\mathbf{z}'_{i,u}: u\in\mathcal{P}(t)\ \wedge\ \mathbf{z}'_{i,u}\ \text{available}\}\right).
\end{equation}
The temporal distance score is then
\begin{equation}
s^{\mathrm{temp}}_{i,t}=d\bigl(\mathbf{z}'_{i,t},\mathbf{B}_{i,t}\bigr),
\end{equation}
where $d(\cdot)$ is either $L_2$ or cosine distance.

\subsubsection{Self-Supervised Change Detection (SSCD)}
SSCD operates on a per-site tensor $\mathbf{Z}_i\in\mathbb{R}^{T\times d}$ and produces three complementary per-month signals:
(a)~\textbf{reconstruction error} $e^{\mathrm{rec}}_{i,t}$ (masked autoencoder),
(b)~\textbf{forecast error} $e^{\mathrm{fore}}_{i,t}$ (next-month predictor), and
(c)~\textbf{latent novelty} $e^{\mathrm{nov}}_{i,t}$ (inverted $k$-NN density in latent space).
After calendar calibration and site-local robust $z$-scoring (yielding normalized signals $z^{\mathrm{rec}}_{i,t}$, $z^{\mathrm{fore}}_{i,t}$, $z^{\mathrm{nov}}_{i,t}$), a weighted ensemble score is computed:
\begin{equation}
b_{i,t}=\alpha_{\mathrm{rec}}\,z^{\mathrm{rec}}_{i,t}+\alpha_{\mathrm{fore}}\,z^{\mathrm{fore}}_{i,t}+\alpha_{\mathrm{nov}}\,z^{\mathrm{nov}}_{i,t}.
\end{equation}
During training, the models are jointly optimized with reconstruction, forecasting, temporal contrastive, and feature decorrelation (Barlow Twins) losses; ensemble weights $\alpha_{\mathrm{rec}}=0.6$, $\alpha_{\mathrm{fore}}=0.3$, $\alpha_{\mathrm{nov}}=0.4$ were selected empirically on the validation split and are applied before per-site min-max normalization to $[0,1]$ (hence their sum need not equal unity).
Both TED and SSCD yield a monthly score per site, mapped to $[0,1]$ via per-site min-max normalization to produce a probability curve $p_{i,t}$.

\subsubsection{Weakly Supervised Monthly Localization}
The WS model trains a temporal classifier to predict the probability of change at each month.
Supervision is weak in three respects: (i)~only a subset of sites have known event months; (ii)~most labels predate 2021, leaving the 2021--2024 period effectively unlabeled (see Fig.~\ref{fig:looted_month}); and (iii)~the labeling protocol assumes a single change event per site, whereas multiple disturbances may occur.
Ground truth provides a binary \texttt{looted} flag and an optional \texttt{looted month}, which is mapped to a change-month index $c_i\in\{0,\dots,T{-}1\}$ when it falls within the 96-month observation window. We define
\begin{equation}
\mathrm{known\_idx}_i=\begin{cases}
 c_i & \text{if looted}=1\ \wedge\ c_i\le c_{\mathrm{end}},\\
 -1  & \text{otherwise},
\end{cases}
\end{equation}
where $c_{\mathrm{end}}$ is the cutoff index corresponding to December 2020.
Training includes all preserved sites and only looted sites with $\mathrm{known\_idx}_i\ge 0$.
A per-month encoder feeds a unidirectional LSTM (2 layers, 128 hidden units, preceded by a 2-layer MLP encoder with LayerNorm projection), followed by a per-time-step head producing month logits $\ell_{i,t}\in\mathbb{R}$ and probabilities
\begin{equation}
p_{i,t}=\sigma(\ell_{i,t})=\frac{1}{1+\exp(-\ell_{i,t})}.
\end{equation}
We train with BCE-with-logits against a smoothed temporal target centered on the labeled month.
Class imbalance is addressed via class weighting and optional oversampling of known-month positive sites.
For preserved sites, the target is $y_{i,t}=0$ for all $t$; looted sites without a known month are excluded from training to prevent incorrect negative labeling of the full timeline.
For looted sites with known change month $c_i$, the target is Gaussian-smoothed: $y_{i,t} = \exp\!\bigl(-(t-c_i)^2 / 2\sigma_w^2\bigr)$ with $\sigma_w=2$ months, providing a soft temporal window around the labeled event.
Optimization uses Adam~\cite{kingma2015adam} with weight decay, gradient clipping, and early stopping on validation loss.

\section{Experimental Setup and Evaluation}
\label{sec:evaluation}

\begin{figure}[!t]
    \centering
    \includegraphics[width=\columnwidth]{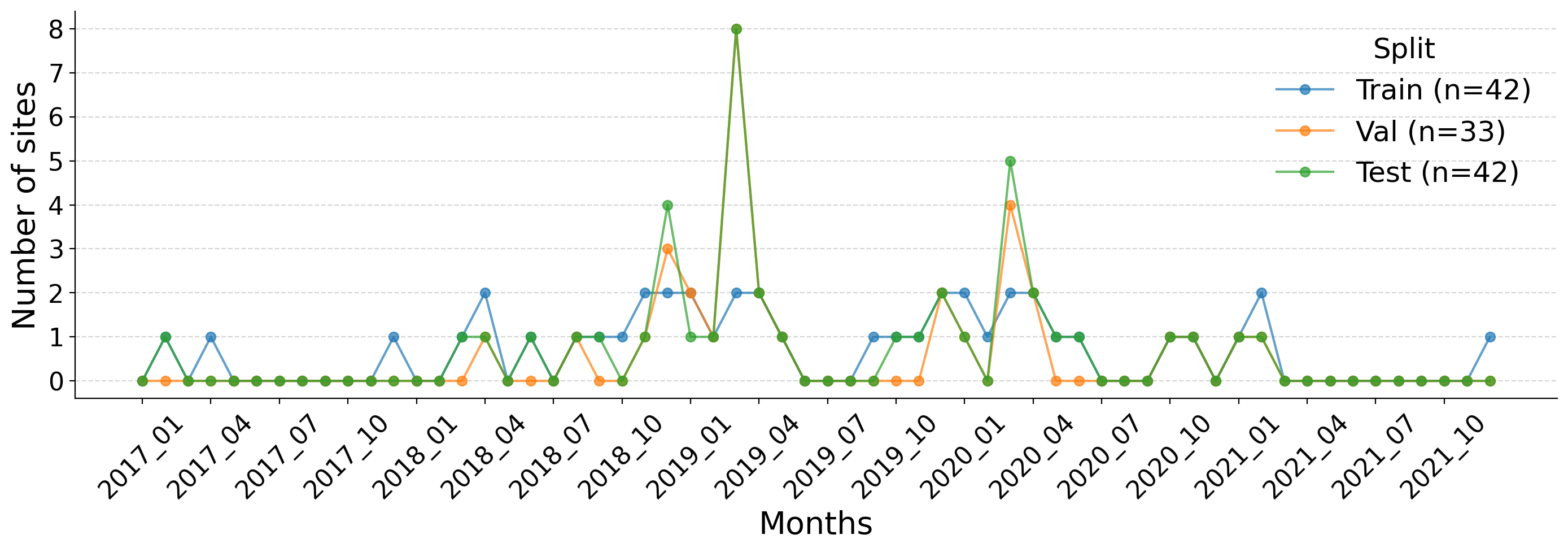}
    \caption{Data splits for the weakly supervised approach.}
    \label{fig:splits}
\end{figure}

Each scoring pipeline produces a probability vector over all 96 months (January 2017 to December 2024) per site, derived by normalizing raw scores (sigmoid or softmax).
For the weakly supervised approach, we use the stratified train/validation/test splits shown in Fig.~\ref{fig:splits}, based on the temporal accumulation of available labels from 2018 to 2020.
All methods are compared on the same test split.
Since ground truth provides at most one known change month for only 117~sites, we evaluate whether the top-$K$ predicted months include the ground-truth event month within a temporal tolerance window.
Of these, 40~sites fall in the test split and 117~comprise the full (``all'') split; all recall percentages should be interpreted accordingly.
Unless otherwise stated, we report results with top-$K=12$ and evaluation restricted to January 2017 to December 2024.

Our evaluation employs a temporally flexible recall metric that accounts for both symmetric and directional detection margins, accommodating the imprecision inherent in recorded looting dates.
For each site $i$ with ground-truth change month $c_i$, let $\{\hat{t}_{ik}\}_{k=1}^K$ denote the indices of the top-$K$ months ranked by $p_{i,t}$ in descending order. We define a symmetric hit indicator, with margin of $m$ month(s):
\begin{equation}
 h_i^{\mathrm{sym}}(m) =
 \begin{cases}
  1, & \text{if } \min_k |\hat{t}_{ik} - c_i| \le m, \\
  0, & \text{otherwise},
 \end{cases}
\end{equation}
and directional indicators that distinguish whether predictions fall after or before the recorded event:
\begin{equation}
 h_i^{+}(m) =
 \begin{cases}
  1, & \text{if } \exists\, k : 0 \le \hat{t}_{ik} - c_i \le m, \\
  0, & \text{otherwise},
 \end{cases}
\end{equation}
\begin{equation}
 h_i^{-}(m) =
 \begin{cases}
  1, & \text{if } \exists\, k : 0 \le c_i - \hat{t}_{ik} \le m, \\
  0, & \text{otherwise}.
 \end{cases}
\end{equation}
Here $h^+_i(m)$ (\emph{positive margin}) fires when at least one predicted month falls \emph{at or after} the ground-truth change month $c_i$ and within $m$ months of it (i.e., $\hat{t}_{ik} \in [c_i,\, c_i{+}m]$), measuring \textbf{confirmation-oriented} recall: the system detects the change only after it has occurred.
For example, if the true looting month is March 2018 and a prediction falls on May 2018, this counts as a positive-margin hit at $m{\ge}2$.
Conversely, $h^-_i(m)$ (\emph{negative margin}) fires when at least one predicted month falls \emph{at or before} $c_i$ and within $m$ months of it (i.e., $\hat{t}_{ik} \in [c_i{-}m,\, c_i]$), measuring \textbf{early-warning} recall: the system flags the site before the recorded disturbance date.
For heritage monitoring, where the operational goal is to detect looting as early as possible to enable timely intervention, high $R^{-}_K(m)$ is the more actionable metric.
Aggregating over $N$ sites with a known month of change, we obtain symmetric and directional temporal recall with margin $m$:
\begin{equation}
 R^{\mathrm{sym}}_K(m) = \frac{1}{N} \sum_{i=1}^N h_i^{\mathrm{sym}}(m),
\end{equation}
\begin{equation}
 R^{+}_K(m) = \frac{1}{N} \sum_{i=1}^N h_i^{+}(m),
\end{equation}
\begin{equation}
 R^{-}_K(m) = \frac{1}{N} \sum_{i=1}^N h_i^{-}(m),
\end{equation}
which jointly quantify temporal accuracy under a tolerance window and expose asymmetries between early and late predictions.

\section{Results}
\label{sec:results}

\subsection{Overall Change Detection Performance}
\begin{figure}[!t]
\centering
\includegraphics[width=\columnwidth]{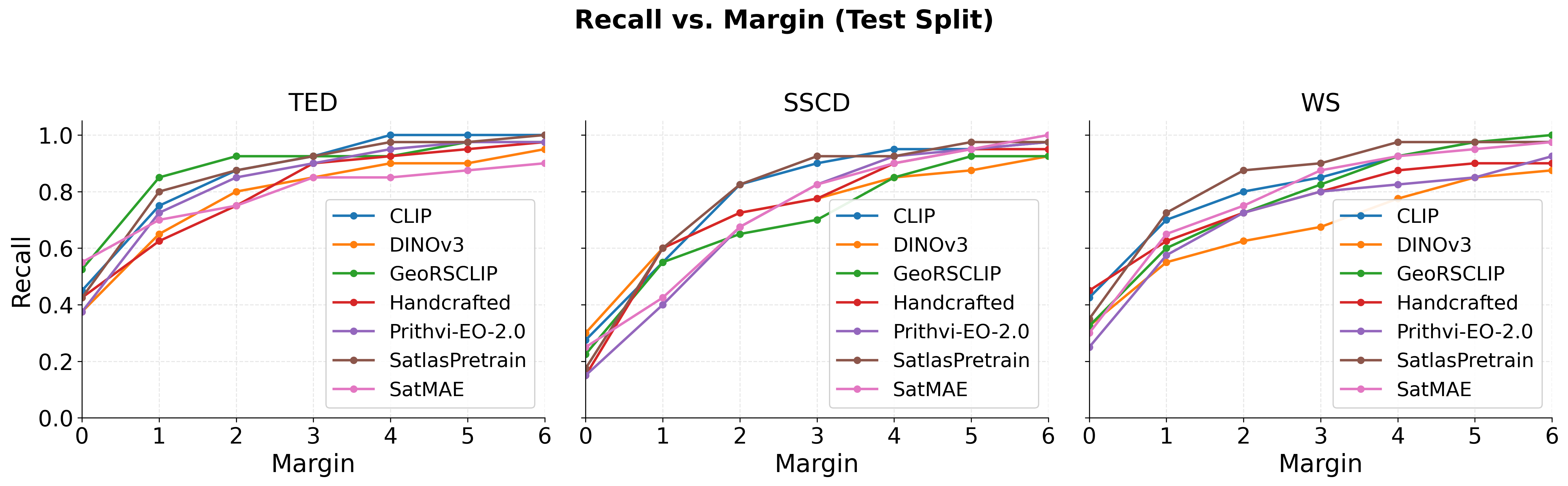}
\caption{Recall (test split) as a function of temporal tolerance $m$ (0--6 months) for all seven embeddings under each scoring approach. Each line represents one embedding; higher curves indicate more robust temporal localization. The steep rise from $m=0$ to $m=3$ confirms that small temporal tolerance substantially increases detection recall across all approaches.}
\label{fig:recall_vs_margin}
\end{figure}

Table~\ref{tab:recall_test_m3} reports recall on the test split for temporal tolerances $m\in\{0,3,6\}$, and Fig.~\ref{fig:recall_vs_margin} shows the full recall curves as a function of margin. Under TED, SatMAE achieves the highest exact-month recall (55.0\% at $m=0$), while GeoRSCLIP, CLIP, and Satlas-Pretrain each reach 92.5\% at the relaxed margin $m=3$. SSCD narrows performance differences at wider margins, with Satlas-Pretrain leading at $m=3$ (92.5\%) and SatMAE reaching 100\% at $m=6$. In the WS setting, Satlas-Pretrain achieves the best $m=3$ recall (90.0\%) and GeoRSCLIP reaches 100\% at $m=6$. Across all approaches, performance at strict $m=0$ varies substantially (15\%--55\%), highlighting intrinsic differences in embedding sensitivity to exact-month change detection.

\begin{table*}[htbp]
\centering
\caption{Recall (\%) on the test split at representative temporal margins. Columns show symmetric margins $m\in\{0,3,6\}$.}
\label{tab:recall_test_m3}
\small
\rowcolors{6}{gray!12}{white}
\begin{tabular}{lccccccccc}
\toprule
& \multicolumn{3}{c}{\textbf{TED}} & \multicolumn{3}{c}{\textbf{SSCD}} & \multicolumn{3}{c}{\textbf{WS}} \\ \cmidrule(r){2-4} \cmidrule(lr){5-7} \cmidrule(l){8-10}
\textbf{Embedding} & $\mathbf{m{=}0}$ & $\mathbf{m{=}3}$ &  $\mathbf{m{=}6}$ &  $\mathbf{m{=}0}$ &  $\mathbf{m{=}3}$ & $\mathbf{m{=}6}$ & $\mathbf{m{=}0}$ &  $\mathbf{m{=}3}$ &  $\mathbf{m{=}6}$ \\
\midrule
CLIP            & 45.0 & \textbf{92.5} & \textbf{100.0} & 27.5 & 90.0 &  97.5 & 42.5 & 85.0 &  97.5 \\
DINOv3          & 37.5 & 85.0 &  95.0 & \textbf{30.0} & 77.5 &  92.5 & 32.5 & 67.5 &  87.5 \\
GeoRSCLIP       & 52.5 & 92.5 &  97.5 & 22.5 & 70.0 &  92.5 & 32.5 & 82.5 & \textbf{100.0} \\
Prithvi-EO-2.0  & 37.5 & 90.0 &  97.5 & 15.0 & 82.5 &  97.5 & 25.0 & 80.0 &  92.5 \\
Satlas-Pretrain & 42.5 & \textbf{92.5} & \textbf{100.0} & 17.5 & \textbf{92.5} &  97.5 & 35.0 & \textbf{90.0} &  97.5 \\
SatMAE          & \textbf{55.0} & 85.0 &  90.0 & 25.0 & 82.5 & \textbf{100.0} & 30.0 & 87.5 &  97.5 \\
Handcrafted     & 42.5 & 90.0 &  97.5 & 15.0 & 77.5 &  95.0 & \textbf{45.0} & 80.0 &  90.0 \\
\bottomrule
\end{tabular}
\end{table*}

\begin{table*}[htbp]
\centering
\caption{Handcrafted features vs.\ best foundation-model embedding per approach on the test split (recall, \%). The best embedding is selected by highest average recall across all margins.}
\label{tab:handcrafted_vs_best}
\small
\begin{tabular}{llccccccc}
\toprule
& & \multicolumn{7}{c}{\textbf{Temporal margin} $\mathbf{m}$} \\ \cmidrule(l){3-9}
\textbf{Approach} & \textbf{Embedding} & $\mathbf{0}$ & $\mathbf{1}$ & $\mathbf{2}$ & $\mathbf{3}$ & $\mathbf{4}$ & $\mathbf{5}$ & $\mathbf{6}$ \\
\midrule
\multirow{2}{*}{TED} & Handcrafted          & 42.5 & 62.5 & 75.0 & 90.0 & 92.5 & 95.0 & 97.5 \\
 & GeoRSCLIP            & \textbf{52.5} & \textbf{85.0} & \textbf{92.5} & \textbf{92.5} & 92.5 & \textbf{97.5} & 97.5 \\
\midrule
\multirow{2}{*}{SSCD}
 & Handcrafted          & 15.0 & \textbf{60.0} & 72.5 & 77.5 & 90.0 & 95.0 & 95.0 \\
& CLIP                 & \textbf{27.5} & 55.0 & \textbf{82.5} & \textbf{90.0} & \textbf{95.0} & 95.0 & \textbf{97.5} \\
\midrule
\multirow{2}{*}{WS}
 & Handcrafted          & \textbf{45.0} & 62.5 & 72.5 & 80.0 & 87.5 & 90.0 & 90.0 \\
& Satlas-Pretrain      & 35.0 & \textbf{72.5} & \textbf{87.5} & \textbf{90.0} & \textbf{97.5} & \textbf{97.5} & \textbf{97.5} \\
\bottomrule
\end{tabular}
\end{table*}

\begin{table*}[htbp]
\centering
\caption{Best-performing embedding per method on the test split (recall, \%).}
\label{tab:best_test_m3}
\small
\rowcolors{5}{gray!12}{white}
\resizebox{\linewidth}{!}{
\begin{tabular}{lcccc}
\toprule
\textbf{Method} & $\mathbf{m{=}0}$ & $\mathbf{m{=}1}$ & $\mathbf{m{=}2}$ & $\mathbf{m{=}3}$ \\
\midrule
TED
& SatMAE (55.0)
& GeoRSCLIP (85.0)
& GeoRSCLIP (92.5)
& CLIP, GeoRSCLIP, Satlas-Pr.\ (92.5) \\

SSCD
& DINOv3 (30.0)
& DINOv3, Handcr., Satlas-Pr.\ (60.0)
& CLIP, Satlas-Pr.\ (82.5)
& Satlas-Pretrain (92.5) \\

WS
& Handcrafted (45.0)
& Satlas-Pr.\ (72.5)
& Satlas-Pr.\ (87.5)
& Satlas-Pr.\ (90.0) \\

\bottomrule
\end{tabular}
}
\end{table*}

\subsection{Foundation-Model Embeddings vs.\ Handcrafted Features}

Table~\ref{tab:handcrafted_vs_best} shows the comparison between handcrafted features and the best foundation-model embedding for each approach across all temporal margins.
Three findings stand out.
First, foundation-model embeddings consistently outperform handcrafted features at strict margins: GeoRSCLIP exceeds Handcrafted by 10 percentage points at $m{=}0$ under TED (52.5\% vs.\ 42.5\%), and CLIP nearly doubles Handcrafted recall under SSCD (27.5\% vs.\ 15.0\%).
Second, the gap narrows or reverses at wider tolerances; under WS, Handcrafted achieves the highest exact-month recall (45.0\% at $m{=}0$), although Satlas-Pretrain overtakes it from $m{\geq}1$ onward and gains 7.5 points by $m{=}6$ (97.5\% vs.\ 90.0\%).
Third, foundation-model embeddings generally saturate faster: GeoRSCLIP reaches 92.5\% by $m{=}2$ under TED, whereas Handcrafted requires $m{=}4$ to reach the same level.
Overall, foundation-model embeddings provide stronger signals for month-level localization, but handcrafted features remain competitive under certain scoring regimes, particularly at exact-month detection in the weakly supervised setting (see also Fig.~\ref{suppfig:recall_by_embedding}).

\subsection{Effect of Weak Supervision}

\begin{figure}[htbp]
\centering
\includegraphics[width=\columnwidth]{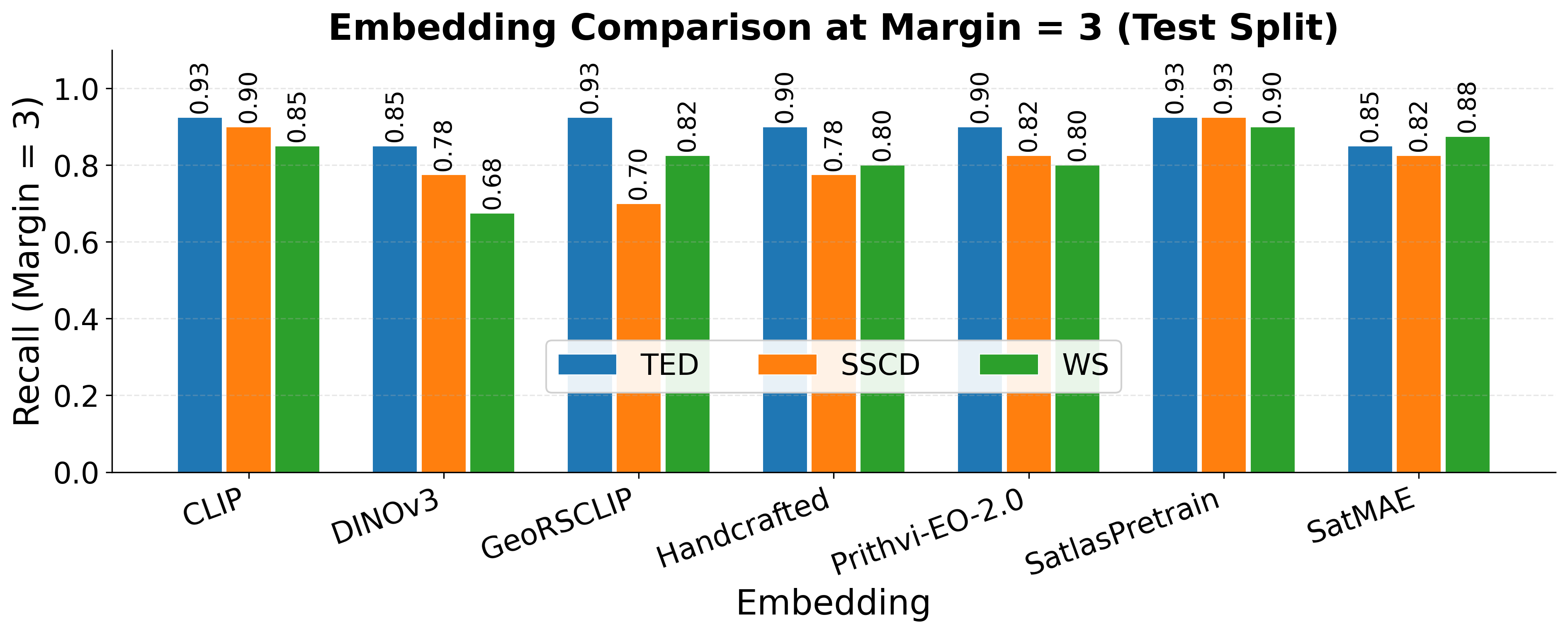}
\caption{Recall at $m=3$ (test split) grouped by embedding, colored by scoring approach. Satlas-Pretrain and GeoRSCLIP consistently lead across approaches (see Table~\ref{tab:best_test_m3} in the main text), while DINOv3 and Handcrafted show more method-dependent variation.}
\label{suppfig:recall_by_embedding}
\end{figure}

Both Table~\ref{tab:recall_test_m3} and Fig.~\ref{suppfig:recall_by_embedding} show that WS generally underperforms the unsupervised methods (TED and SSCD).
While weak supervision can sharpen localization when sufficient known-month labels are available, it tends to yield weaker results when labels are sparse or when the learned decision boundary does not generalize across embeddings.
In our setup, the weakly supervised model is competitive for some embeddings (e.g., GeoRSCLIP and SatMAE at $m=3$), but the best-performing configurations overall are achieved by unsupervised pipelines.
Table~\ref{tab:best_test_m3} provides a compact view of the best embedding per method at each symmetric margin $m\in\{0,\ldots,3\}$.

\subsection{Unsupervised Scoring Approaches: TED vs.\ SSCD}

TED offers a simple, training-free scoring rule, making it attractive when computational resources or training data are limited.
SSCD adds modeling capacity through reconstruction, forecasting, and novelty estimation.
The macro-averaged gap between TED and SSCD narrows monotonically as temporal tolerance widens (Table~\ref{tab:dist_vs_unsup_macro_m0_to_m6_nodino}): TED leads by 18.4 points at $m=0$ but the two methods reach near-parity at $m=6$ ($\Delta = +0.1$).
Per-embedding wins for SSCD remain limited at stricter margins.
These results indicate that TED remains a strong, no-training baseline, whereas SSCD becomes competitive only at wider temporal tolerances.

\begin{table}[!t]
\centering
\caption{\textbf{TED vs.\ SSCD.}
Macro-averaged recall (\%) on the full split at symmetric temporal margins $m\in\{0,\ldots,6\}$.
$\Delta$ denotes the mean improvement (SSCD minus TED).}
\label{tab:dist_vs_unsup_macro_m0_to_m6_nodino}
\small
\rowcolors{4}{gray!12}{white}
\begin{tabular}{lccc}
\toprule
&\multicolumn{2}{c}{\textbf{Method}}  & \\ \cmidrule{2-3}
\textbf{Margin} & \textbf{TED} & \textbf{SSCD} & $\boldsymbol{\Delta}$ \textbf{mean} \\
\midrule
$m=0$ & 40.1 & 21.8 & $-18.4$ \\
$m=1$ & 67.8 & 54.1 & $-13.6$ \\
$m=2$ & 81.7 & 73.1 & $-8.5$ \\
$m=3$ & 89.0 & 82.4 & $-6.5$ \\
$m=4$ & 94.0 & 90.6 & $-3.4$ \\
$m=5$ & 95.2 & 94.4 & $-0.8$ \\
$m=6$ & 97.0 & 97.1 & $+0.1$ \\
\bottomrule
\end{tabular}
\end{table}

\subsection{Directional Margin Asymmetry}

To assess whether methods tend to detect changes before or after the recorded event month, we analyze directional margin asymmetry.
For each embedding and scoring method, we compute the average gap $\bar{\Delta} = \frac{1}{7}\sum_{m=0}^{6}\bigl[R^{+}_K(m) - R^{-}_K(m)\bigr]$, where $R^{+}_K$ is the positive-margin (post-event) recall and $R^{-}_K$ is the negative-margin (pre-event) recall.
A \textbf{negative gap} indicates that the method more often flags change \emph{before} the recorded looting month (early-warning bias), which is operationally preferable for timely intervention.
A \textbf{positive gap} indicates that the method more often flags change \emph{after} the recorded month (confirmation bias), which is useful but less actionable for prevention.

\begin{figure}[!t]
\centering
\includegraphics[width=\columnwidth]{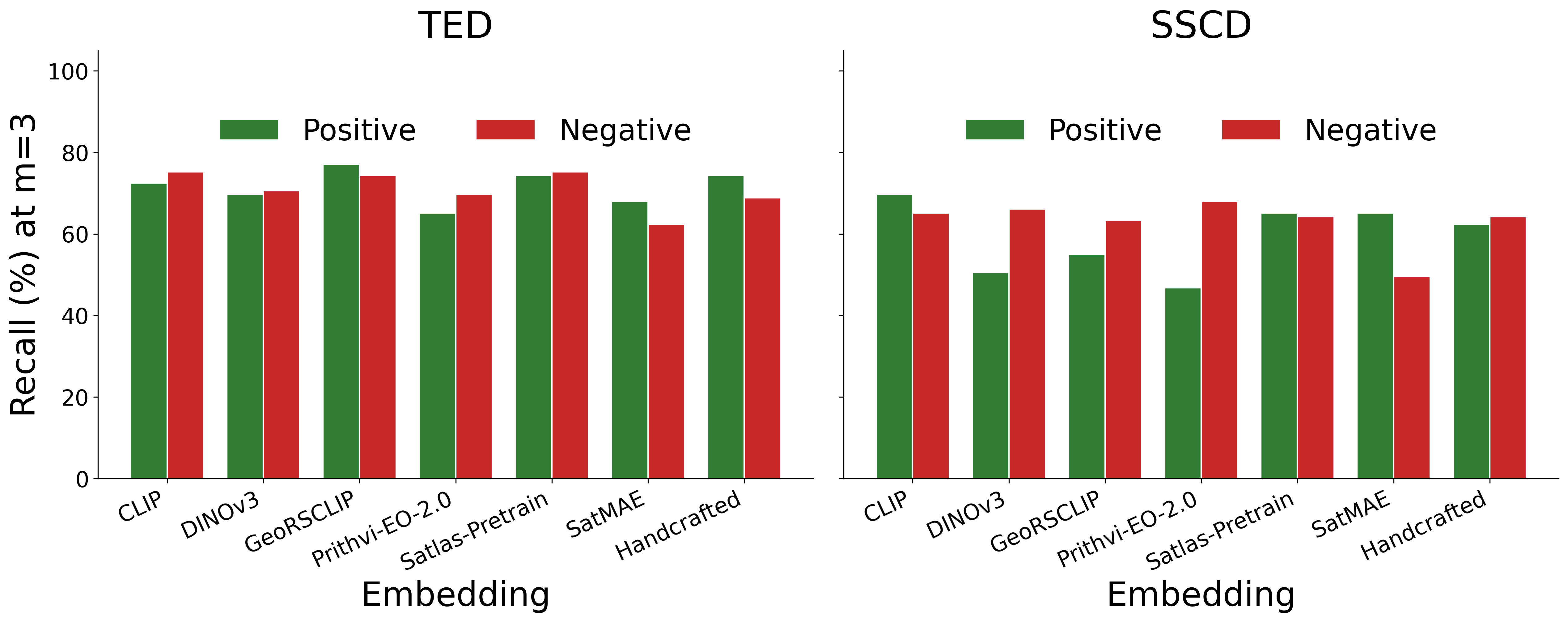}
\caption{Directional recall at $m=3$ (all split), showing positive (late-detection) versus negative (early-detection) margin recall for each embedding and approach.}
\label{suppfig:directional_all}
\end{figure}

Table~\ref{tab:directional_gap_test_avg} reports $\bar{\Delta}$ for both the test and all splits.
On the test split, SSCD exhibits the strongest early-warning tendency, with large negative gaps for GeoRSCLIP ($-15.4$), DINOv3 ($-11.8$), and Prithvi-EO-2.0 ($-11.1$), meaning these embedding--method combinations frequently detect anomalies in the months leading up to the recorded event.
TED, by contrast, leans toward confirmation for most embeddings (5/7 positive gaps), consistent with its design: comparing against preceding months naturally produces elevated scores \emph{after} a change has occurred.
On the all split, directional biases generally attenuate, indicating that extreme asymmetries observed on the test split are partly due to the small sample size.
However, SSCD retains a consistent early-warning advantage for DINOv3 ($-9.3$), Prithvi-EO-2.0 ($-9.2$), and GeoRSCLIP ($-7.6$), confirming that the reconstruction and forecasting components of SSCD are genuinely sensitive to precursor signals (see Fig.~\ref{suppfig:directional_all} for a visual comparison at $m=3$).
For practitioners prioritizing early detection, pairing SSCD with GeoRSCLIP or Prithvi-EO-2.0 embeddings offers the most favorable early-warning profile across both evaluation splits.

\begin{table}[!t]
\centering
\caption{Directional margin asymmetry averaged across all margins ($m=0\ldots6$).
Values report the average positive-margin recall minus negative-margin recall (percentage points).}
\label{tab:directional_gap_test_avg}
\small
\rowcolors{4}{gray!12}{white}
\resizebox{\columnwidth}{!}{
\begin{tabular}{lcccccc}
\toprule
& \multicolumn{3}{c}{\textbf{Test split}} & \multicolumn{3}{c}{\textbf{All split}} \\ \cmidrule(r){2-4} \cmidrule(l){5-7}
\textbf{Embedding} & \textbf{TED} & \textbf{SSCD} & \textbf{WS} & \textbf{TED} & \textbf{SSCD} & \textbf{WS} \\
\midrule
CLIP                & $-3.9$  & $-5.7$  & $-2.9$ & $-2.3$ & $+5.0$  & $-7.5$ \\
DINOv3              & $+5.4$  & $-11.8$ & $-1.8$ & $+0.1$ & $-9.3$  & $+0.5$ \\
GeoRSCLIP           & $+1.1$  & $-15.4$ & $+7.9$ & $+3.7$ & $-7.6$  & $+0.9$ \\
Prithvi-EO-2.0      & $-2.5$  & $-11.1$ & $+12.9$ & $-5.5$ & $-9.2$ & $+1.3$ \\
Satlas-Pretrain     & $+6.8$  & $-1.4$  & $+0.7$ & $+0.7$ & $+0.7$  & $+1.7$ \\
SatMAE              & $+3.2$  & $+6.4$  & $-5.4$ & $+3.5$ & $+8.5$  & $-1.7$ \\
Handcrafted         & $+11.1$ & $+2.1$  & $-6.8$ & $+1.7$ & $+1.7$  & $-2.2$ \\
\bottomrule
\end{tabular}
}
\end{table}

\subsection{Per-Site Score Trajectories and Feature-Space Variability}

Per-site monthly score trajectories (Fig.~\ref{suppfig:site_examples}) illustrate the complementary strengths of the three approaches: TED and SSCD offer broad temporal sensitivity with peaks near recorded looting events, while WS produces sharp spikes only when the event falls within the temporal range of training labels.
An analysis of feature-space variability over time (Fig.~\ref{suppfig:feature_variation}) reveals three distinct tiers of cross-site variability: CLIP (highest), DINOv3, GeoRSCLIP, and SatMAE (middle), and Prithvi-EO-2.0, Satlas-Pretrain, and Handcrafted (lowest).

\section{Change Detection Across Global Sites}
\label{sec:global}

\begin{figure*}[!t]
    \centering
    \includegraphics[width=\textwidth]{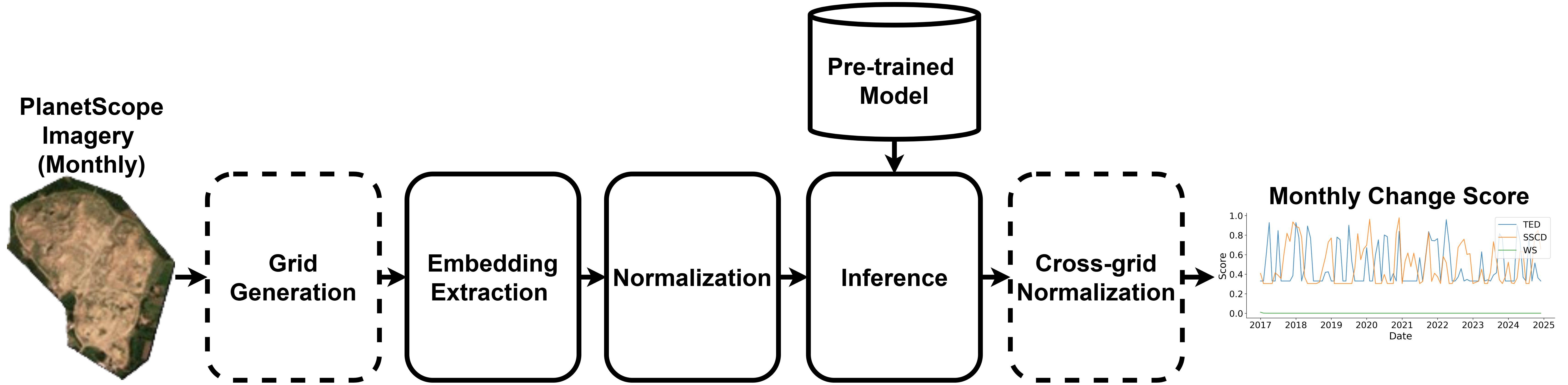}
    \caption{Block diagram of the WATCH global site inference pipeline. Monthly PlanetScope mosaics are gridded into $1\,\mathrm{km}^2$ tiles, encoded by a frozen foundation model, and normalized using Afghanistan-derived statistics. The normalized embeddings are then routed to all three scoring branches (TED, SSCD, WS) in parallel. Scores are cross-grid normalized to suppress uniform seasonal signals, then aggregated per site via max or mean pooling.}
    \label{fig:global_sites}
\end{figure*}

\begin{figure*}[!t]
    \centering
    \begin{subfigure}[b]{0.49\textwidth}
        \centering
        \includegraphics[width=\textwidth]{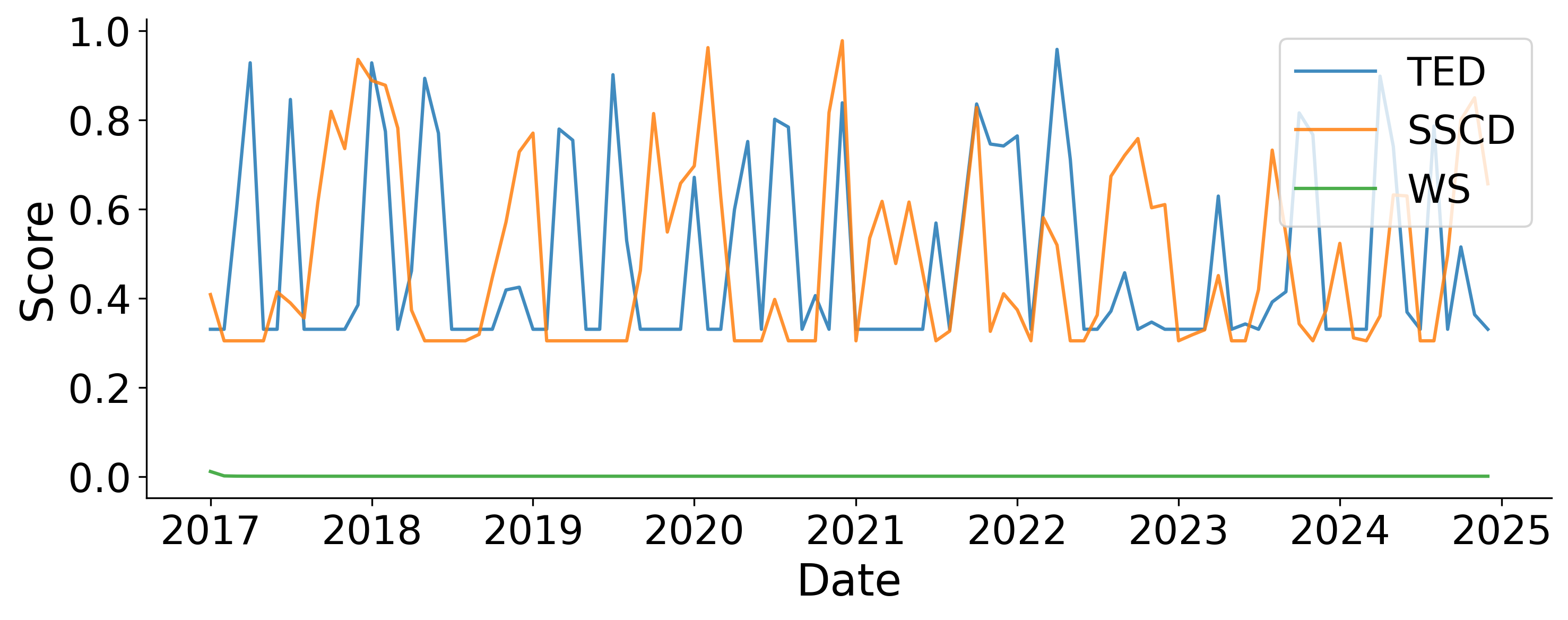}
    \caption{Egypt: Lisht (grid 7)}
    \end{subfigure}
    \hfill
    \begin{subfigure}[b]{0.49\textwidth}
        \centering
        \includegraphics[width=\textwidth]{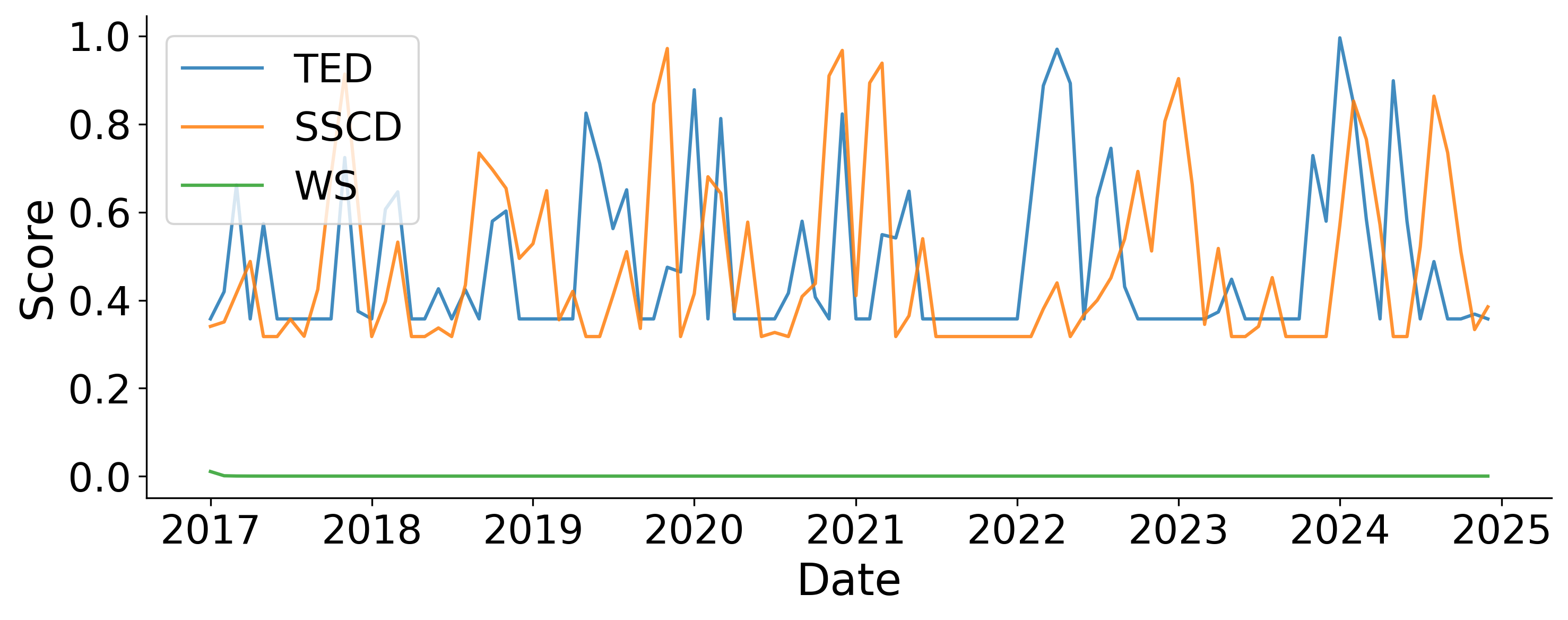}
        \caption{Pakistan: Charsadda SW (grid 0)}
    \end{subfigure}
    \\[0.5em]
    \begin{subfigure}[b]{0.49\textwidth}
        \centering
        \includegraphics[width=\textwidth]{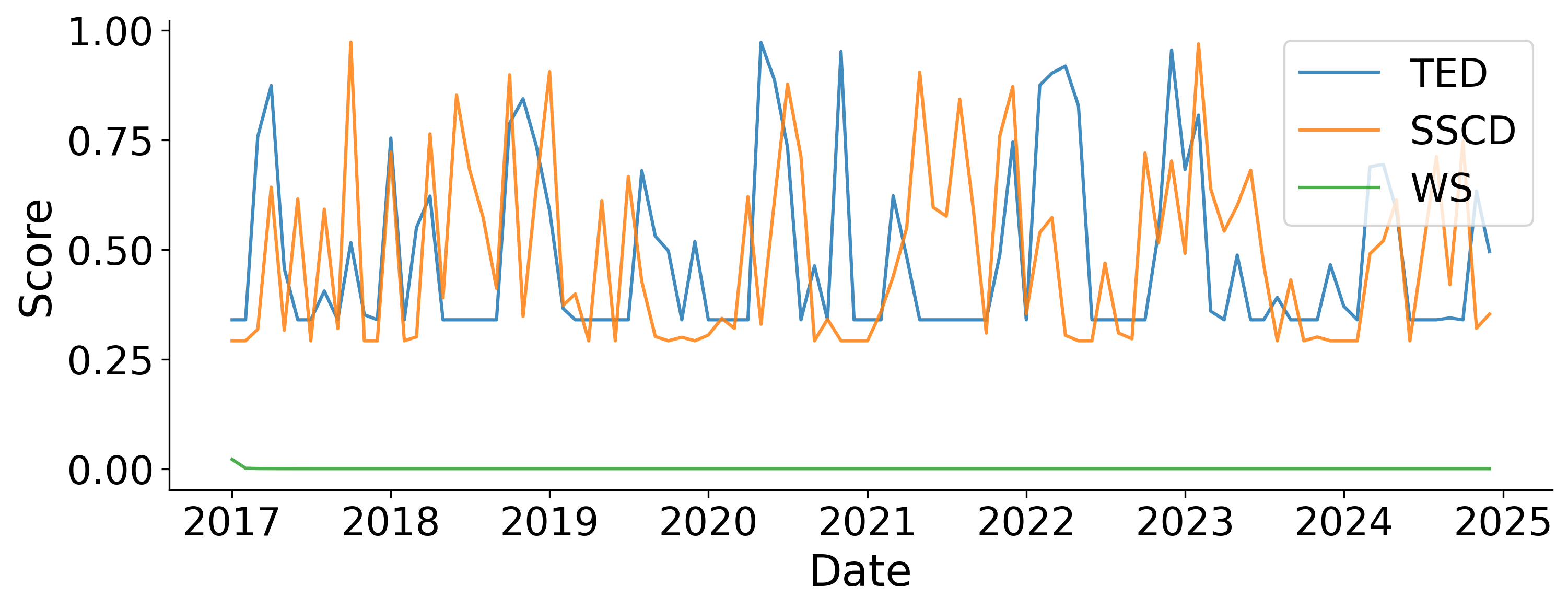}
        \caption{Syria: Dura Europos (grid 0)}
    \end{subfigure}
    \hfill
    \begin{subfigure}[b]{0.49\textwidth}
        \centering
        \includegraphics[width=\textwidth]{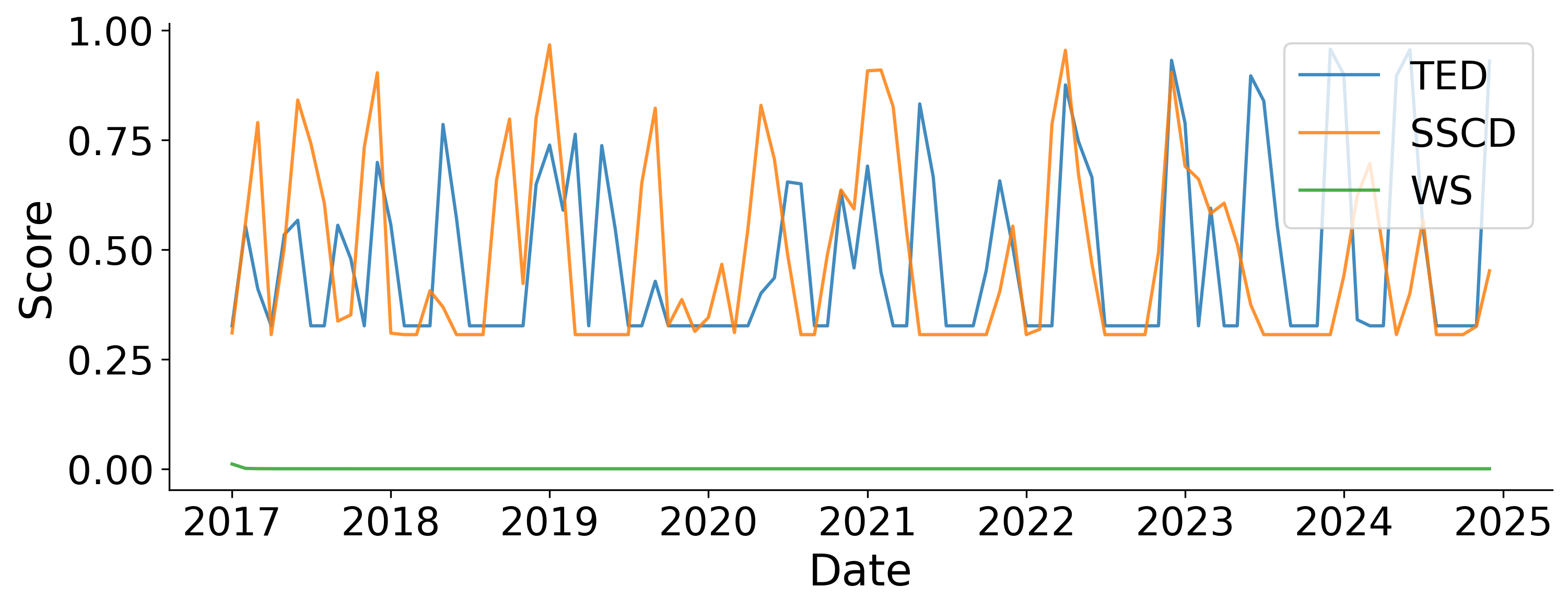}
        \caption{Turkey: Kaymak (grid 6)}
    \end{subfigure}
    \caption{Monthly GeoRSCLIP scores under three scoring regimes (TED, SSCD, WS) for representative grid cells at four global heritage sites. TED and SSCD transfer effectively without retraining, producing temporally structured score trajectories, while WS remains near-zero outside the Afghan training domain.}
    \label{fig:global_modes_all_sites}
\end{figure*}

We apply the WATCH framework to sites in Turkey, Syria, Pakistan, and Egypt to assess cross-regional transfer (see Fig.~\ref{fig:global_sites} for the global inference pipeline).
Because no ground-truth event months are available for these sites, this constitutes a qualitative operational demonstration rather than a quantitative evaluation.
Feature normalization parameters and learned model weights are transferred directly from the Afghanistan setup without re-estimation; when a site spans multiple grid cells, cross-grid normalization suppresses uniform seasonal signals and highlights localized changes.
Fig.~\ref{fig:global_modes_all_sites} shows GeoRSCLIP monthly scores for representative grid cells at four global sites: TED and SSCD produce temporally rich score trajectories that transfer effectively without retraining, while WS scores remain near-zero, reflecting the fact that the weakly supervised model was trained exclusively on Afghan sites whose temporal label distribution does not transfer to other regions.
This contrast underscores a key practical finding: unsupervised approaches generalize to out-of-distribution sites without retraining, whereas WS requires further calibration using region-specific labeled data.

\section{Discussion}
\label{sec:discussion}

\subsection{Representation Choice: Foundation Models vs.\ Handcrafted Features}
Geospatial foundation models provide compact representations that reduce sensitivity but remaining responsive to localized structural or textural changes.
Across multiple embeddings, these representations enable effective month-level localization even with simple unsupervised scoring rules.
However, performance varies substantially across embedding families, indicating that pretraining objectives and inductive biases strongly influence downstream temporal behavior.
Embeddings trained with explicit geospatial or multimodal alignment objectives (e.g., GeoRSCLIP, Prithvi-EO-2.0) tend to yield stronger temporal signals than generic visual pretraining.
At the same time, handcrafted spectral and texture features remain competitive and, in some configurations, superior.
These findings indicate that hybrid systems combining both representation types remain a viable design choice.

\subsection{Supervision Trade-Offs}
The consistent underperformance of WS relative to unsupervised methods (Table~\ref{tab:recall_test_m3}) likely reflects a structural limitation rather than a modeling failure: with only $\sim$60 known-month labels concentrated before 2021, the LSTM learns temporal priors tied to the label distribution rather than generalizable change signatures.
This suggests that weak supervision for temporal localization requires either substantially more labeled events or explicit data-augmentation strategies to prevent temporal overfitting.
In contrast, TED's training-free design and SSCD's self-supervised objectives avoid this requirement entirely, making them preferable defaults when labeled data is scarce.

\subsection{Directional Temporal Asymmetry}
Directional margin analysis goes beyond aggregate recall by revealing whether methods tend to flag changes before or after the recorded event month.
For operational heritage monitoring, early-warning capability (negative margin recall, $R^{-}_K$) is critical: detecting looting before or at the reported month enables timely field responses and potential intervention.
Our results show that SSCD paired with geospatial foundation-model embeddings (GeoRSCLIP, Prithvi-EO-2.0, DINOv3) consistently achieves the strongest early-warning profile, likely because the reconstruction and forecasting components detect subtle precursor anomalies before the labeled event.
TED, by contrast, inherently favors confirmation-oriented detection, since it scores deviation from preceding months and thus reacts after a change has materialized.
These findings provide actionable guidance: practitioners seeking early warning should favor SSCD, while those requiring post-event forensic confirmation can rely on TED.

\subsection{Transfer to Global Sites}
Applying WATCH to sites outside Afghanistan demonstrates that the framework can generalize across regions using frozen embeddings and calibration parameters learned from a single country, particularly the unsupervised change scoring approaches.
While this transfer carries the risk of distribution shift, the qualitative results across Syria, Turkey, Pakistan, and Egypt suggest that the learned representations capture broadly reusable temporal signals.
The grid-based formulation further enables localized change detection within larger sites, supporting practical deployment for wide-area heritage monitoring.

\section{Limitations}
\label{sec:limitations}

First, reliable month-level ground truth is extremely scarce, with only a small fraction of looted sites having known disturbance dates, which limits both supervised learning and evaluation of early-warning capabilities.
Second, the assumption of a single dominant change event per site simplifies modeling but does not capture repeated or prolonged disturbance patterns.
Third, global transfer relies on normalization and model parameters learned from Afghanistan, which may introduce distribution shift in regions with different land cover, climate, or imaging conditions.
Finally, monthly Planet mosaics, while operationally convenient, may miss short-lived or rapidly evolving events that occur between acquisition intervals.

\section{Conclusion}
\label{sec:conclusion}

We introduced WATCH, a framework for monthly change localization in archaeological site monitoring that operates under sparse supervision and at large scale.
By combining site-centric time series, geospatial foundation-model embeddings, and three complementary scoring approaches (TED, SSCD, and WS), WATCH enables interpretable month-level change detection with minimal annotation requirements.
The best unsupervised configuration (TED with SatMAE) achieves 55\% exact-month recall at $m{=}0$, while TED with GeoRSCLIP reaches 92.5\% within a 3-month tolerance.

Our directional margin analysis yields concrete operational guidance: practitioners seeking \emph{early warning} should pair SSCD with GeoRSCLIP or Prithvi-EO-2.0 embeddings, which consistently exhibit the strongest pre-event detection profiles, while those requiring \emph{post-event forensic confirmation} can rely on TED.
Handcrafted features retain competitive value, particularly for exact-month detection under weak supervision.

The most pressing avenue for future work is expanding ground-truth coverage, currently limited to 117 sites with known disturbance months, e.g., through active learning, which would strengthen both supervised training and evaluation reliability.
WATCH bridges recent advances in geospatial representation learning with the operational realities of cultural heritage protection, supporting scalable and globally applicable monitoring systems.

\bibliographystyle{IEEEtran}
\bibliography{references}

@InProceedings{tadesse2026satellite,
    author    = {Tadesse, Girmaw Abebe and Bartette, Titien and Hassanali, Andrew and Kim, Allen and Chemla, Jonathan and Zolli, Andrew and Ubelmann, Yves and Robinson, Caleb and Becker-Reshef, Inbal and Ferres, Juan Lavista},
    title     = {Satellite-Based Detection of Looted Archaeological Sites Using Machine Learning},
    booktitle = {Proceedings of the IEEE/CVF Winter Conference on Applications of Computer Vision (WACV) Workshops},
    month     = {March},
    year      = {2026},
    pages     = {840-848}
}

@inproceedings{radford2021learning,
  title={Learning transferable visual models from natural language supervision},
  author={Radford, Alec and Kim, Jong Wook and Hallacy, Chris and Ramesh, Aditya and Goh, Gabriel and Agarwal, Sandhini and Sastry, Girish and Askell, Amanda and Mishkin, Pamela and Clark, Jack and others},
  booktitle={International Conference on Machine Learning (ICML)},
  pages={8748--8763},
  year={2021}
}

@misc{simeoni2025dinov3,
  title={{DINO}v3},
  author={Oriane Siméoni and Huy V. Vo and Maximilian Seitzer and Federico Baldassarre and Maxime Oquab and Cijo Jose and Vasil Khalidov and Marc Szafraniec and Seungeun Yi and Michaël Ramamonjisoa and Francisco Massa and Daniel Haziza and Luca Wehrstedt and Jianyuan Wang and Timothée Darcet and Théo Moutakanni and Leonel Sentana and Claire Roberts and Andrea Vedaldi and Jamie Tolan and John Brandt and Camille Couprie and Julien Mairal and Hervé Jégou and Patrick Labatut and Piotr Bojanowski},
  year={2025},
  eprint={2508.10104},
  archivePrefix={arXiv}
}

@article{zhang2024georsclip,
  title={{RS}5{M} and {G}eo{R}{S}{C}LIP: A Large Scale Vision-Language Dataset and a Large Vision-Language Model for Remote Sensing},
  author={Zhang, Zilun and Zhao, Tiancheng and Guo, Yulong and Yin, Jianwei},
  journal={IEEE Transactions on Geoscience and Remote Sensing},
  year={2024},
}

@article{szwarcman2025prithvi,
  title={{Prithvi-EO-2.0}: A versatile multi-temporal foundation model for earth observation applications},
  author={Szwarcman, Daniela and Roy, Sujit and Fraccaro, Paolo and G{\'\i}slason, Orsteinn El{\'\i} and Blumenstiel, Benedikt and Ghosal, Rinki and De Oliveira, Pedro Henrique and de Sousa Almeida, Joao Lucas and Sedona, Rocco and Kang, Yanghui and others},
  journal={IEEE Transactions on Geoscience and Remote Sensing},
  year={2025}
}

@book{ball2019archaeological,
  title={Archaeological gazetteer of {A}fghanistan},
  author={Ball, Warwick},
  year={2019},
  publisher={Oxford University Press}
}

@article{menze2012mapping,
  title={Mapping patterns of long-term settlement in {N}orthern {M}esopotamia at a large scale},
  author={Menze, Bjoern H and Ur, Jason A},
  journal={Proceedings of the National Academy of Sciences},
  volume={109},
  number={14},
  pages={E778--E787},
  year={2012}
}

@inproceedings{cong2022satmae,
  author={Cong, Yezhen and Khanna, Samar and Meng, Chenlin and Liu, Patrick and Rozi, Erik and He, Yutong and Burke, Marshall and Lobell, David and Ermon, Stefano},
  title={{S}at{M}AE: Pre-training transformers for temporal and multi-spectral satellite imagery},
  booktitle={Advances in Neural Information Processing Systems (NeurIPS)},
  volume={35},
  pages={197--211},
  year={2022}
}

@misc{jakubik2023foundation,
  author={Jakubik, Johannes and Roy, Sujit and Phillips, C. E. and Fraccaro, Paolo and Godwin, Geoff and Zadrozny, Mirek and Szwarcman, Carlos and Gomes, Srinivasan and Nyirjesy, Sergio and Edwards, Denys and others},
  title={Foundation models for generalist geospatial artificial intelligence},
  howpublished={arXiv:2310.18660},
  year={2023}
}

@inproceedings{bastani2023satlaspretrain,
  author={Bastani, Favyen and Wolters, Piper and Gupta, Ritwik and Ferdinando, Joe and Kembhavi, Aniruddha},
  title={{S}atlas{P}retrain: A large-scale dataset for remote sensing image understanding},
  booktitle={IEEE/CVF International Conference on Computer Vision (ICCV)},
  pages={16772--16782},
  year={2023}
}

@inproceedings{klemmer2024satclip,
  author={Klemmer, Konstantin and Rolf, Esther and Robinson, Caleb and Mackey, Lester and Russwurm, Marc},
  title={{S}at{CLIP}: Global, general-purpose location embeddings with satellite imagery},
  booktitle={AAAI Conference on Artificial Intelligence},
  volume={38},
  number={12},
  pages={13156--13164},
  year={2024}
}

@article{negula2015earth,
  title={Earth observation for the world cultural and natural heritage},
  author={Negula, Iulia Dana and Sofronie, Ramiro and Virsta, Ana and Badea, Alexandru},
  journal={Agriculture and Agricultural Science Procedia},
  volume={6},
  pages={438--445},
  year={2015},
  publisher={Elsevier}
}

@article{cuca2023monitoring,
  title={Monitoring of damages to cultural heritage across {E}urope using remote sensing and earth observation: Assessment of scientific and grey literature},
  author={Cuca, Branka and Zaina, Federico and Tapete, Deodato},
  journal={Remote Sensing},
  volume={15},
  number={15},
  pages={3748},
  year={2023}
}

@inproceedings{vincent2025detecting,
  title={Detecting Looted Archaeological Sites from Satellite Image Time Series},
  author={Vincent, Elliot and Saroufim, Mehra{\"{\i}}l and Chemla, Jonathan and Ubelmann, Yves and Marquis, Philippe and Ponce, Jean and Aubry, Mathieu},
  booktitle={Proceedings of the Computer Vision and Pattern Recognition Conference (CVPR)},
  pages={2296--2307},
  year={2025}
}

@article{parcak2016satellite-evidence,
  title={Satellite evidence of archaeological site looting in {E}gypt: 2002--2013},
  author={Parcak, Sarah and Gathings, David and Childs, Chase and Mumford, Greg and Cline, Eric},
  journal={Antiquity},
  volume={90},
  number={349},
  pages={188--205},
  year={2016}
}

@article{levin2019world,
  title={World {H}eritage in danger: Big data and remote sensing can help protect sites in conflict zones},
  author={Levin, Noam and Ali, Saleem and Crandall, David and Kark, Salit},
  journal={Global Environmental Change},
  volume={55},
  pages={97--104},
  year={2019}
}

@article{tapete2019looting,
  title={Detection of archaeological looting from space: Methods, achievements and challenges},
  author={Tapete, Deodato and Cigna, Francesca},
  journal={Remote Sensing},
  volume={11},
  number={20},
  pages={2389},
  year={2019}
}

@article{agapiou2021unesco,
  title={{UNESCO} World Heritage properties in changing and dynamic environments: change detection methods using optical and radar satellite data},
  author={Agapiou, Athos},
  journal={Heritage Science},
  volume={9},
  number={1},
  pages={1--14},
  year={2021}
}

@article{daudt2018fully,
  title={Fully convolutional siamese networks for change detection},
  author={Daudt, Rodrigo Caye and Le Saux, Bertrand and Boulch, Alexandre},
  journal={International Conference on Image Processing (ICIP)},
  year={2018}
}

@article{chalapathy2019deep,
  title={Deep learning for anomaly detection: A survey},
  author={Chalapathy, Raghavendra and Chawla, Sanjay},
  journal={ACM Computing Surveys},
  volume={54},
  number={3},
  pages={1--38},
  year={2021}
}

@inproceedings{kingma2015adam,
  author={Kingma, Diederik P. and Ba, Jimmy},
  title={{A}dam: A method for stochastic optimization},
  booktitle={International Conference on Learning Representations (ICLR)},
  year={2015}
}


\clearpage
\newpage
\appendices

\section{Dataset: Global Site Map and Per-Country Site Examples}
\label{app:global_sites}

\begin{figure}[htbp]
    \centering
    \includegraphics[width=\columnwidth]{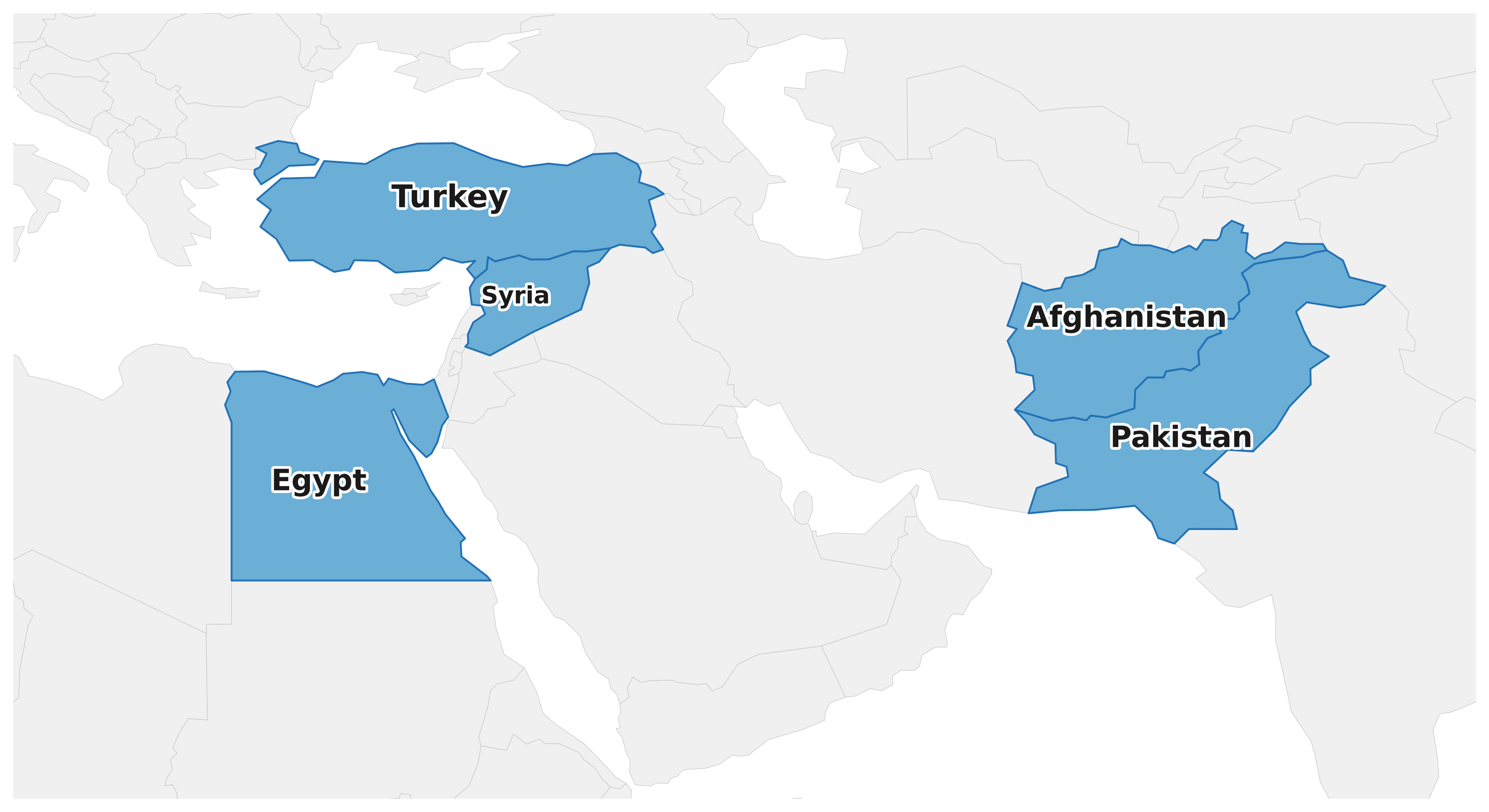}
    \caption{Countries with archaeological sites selected for evaluation in this study to assess the applicability and generalizability of the proposed framework. Per-country site examples are shown in Fig.~\ref{suppfig:global_countries_detail}.}
    \label{suppfig:global_and_countries}
\end{figure}

\begin{figure}[htbp]
    \centering
    \begin{subfigure}[b]{0.15\columnwidth}
        \centering
        \includegraphics[width=\textwidth]{figures/afghanistan/temporal_sites/looted_171/2022_12.png}
        \caption{Afghanistan}
    \end{subfigure}
    \begin{subfigure}[b]{0.19\columnwidth}
        \centering
        \includegraphics[width=\textwidth]{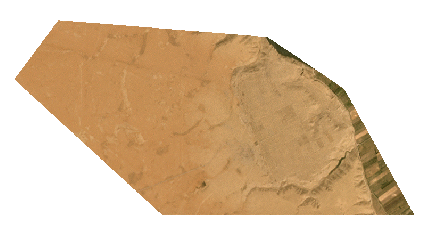}
        \caption{Syria}
    \end{subfigure}
    \begin{subfigure}[b]{0.19\columnwidth}
        \centering
        \includegraphics[width=\textwidth]{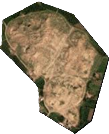}
        \caption{Pakistan}
    \end{subfigure}
    \begin{subfigure}[b]{0.19\columnwidth}
        \centering
        \includegraphics[width=\textwidth]{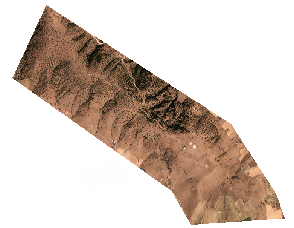}
        \caption{Turkey}
    \end{subfigure}
    \begin{subfigure}[b]{0.19\columnwidth}
        \centering
        \includegraphics[width=\textwidth]{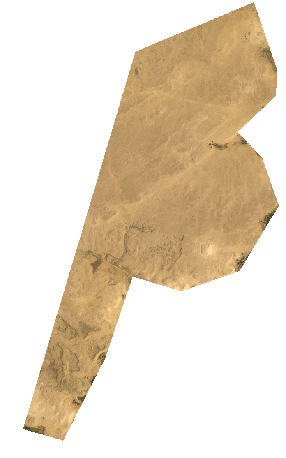}
        \caption{Egypt}
    \end{subfigure}
    \caption{Examples of PlanetScope monthly mosaics for archaeological sites in five countries evaluated in this study.}
    \label{suppfig:global_countries_detail}
\end{figure}

\section{Additional Results Figures}
\label{app:additional_results}

\section{Per-Site Score Trajectories}
\label{app:trajectories}

\begin{figure*}[htbp]
    \centering
    \begin{subfigure}[b]{0.45\textwidth}
        \centering
        \includegraphics[width=\textwidth]{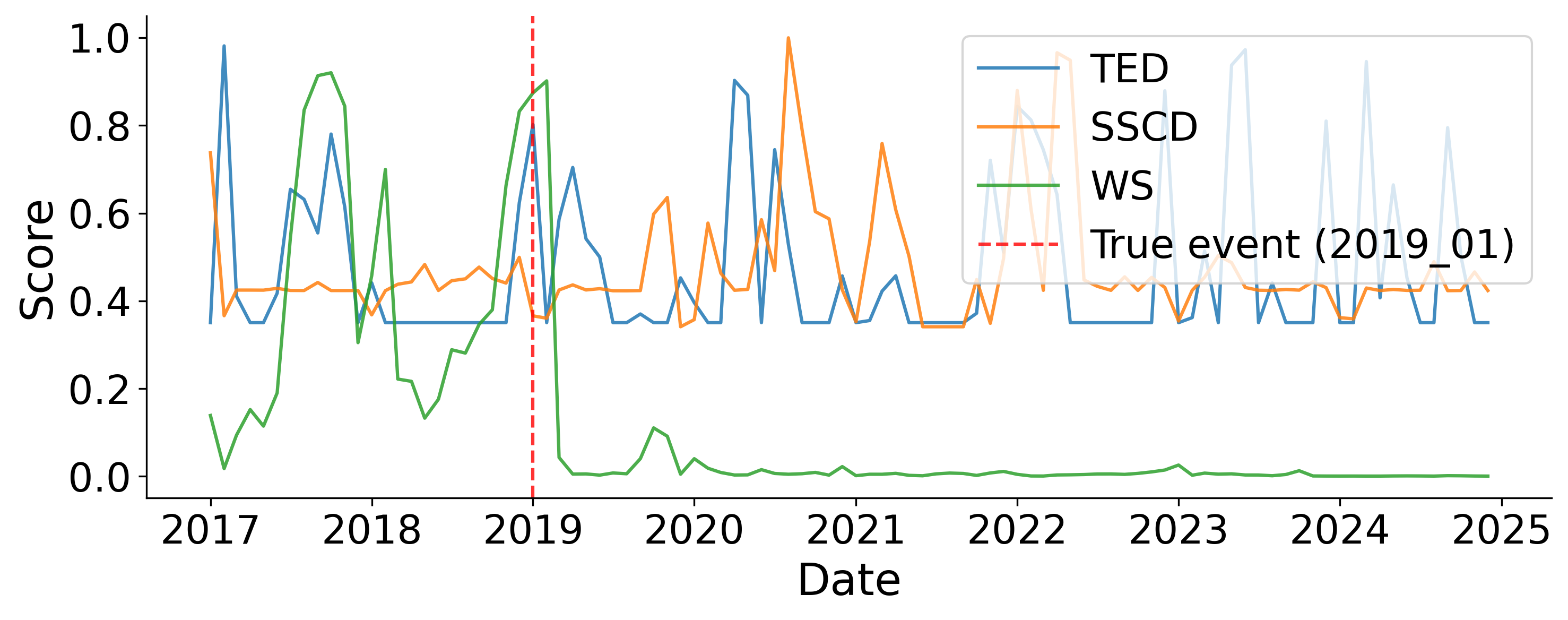}
        \caption{Site looted\_799 (event: January 2019)}
    \end{subfigure}
    \hfill
    \begin{subfigure}[b]{0.45\textwidth}
        \centering
        \includegraphics[width=\textwidth]{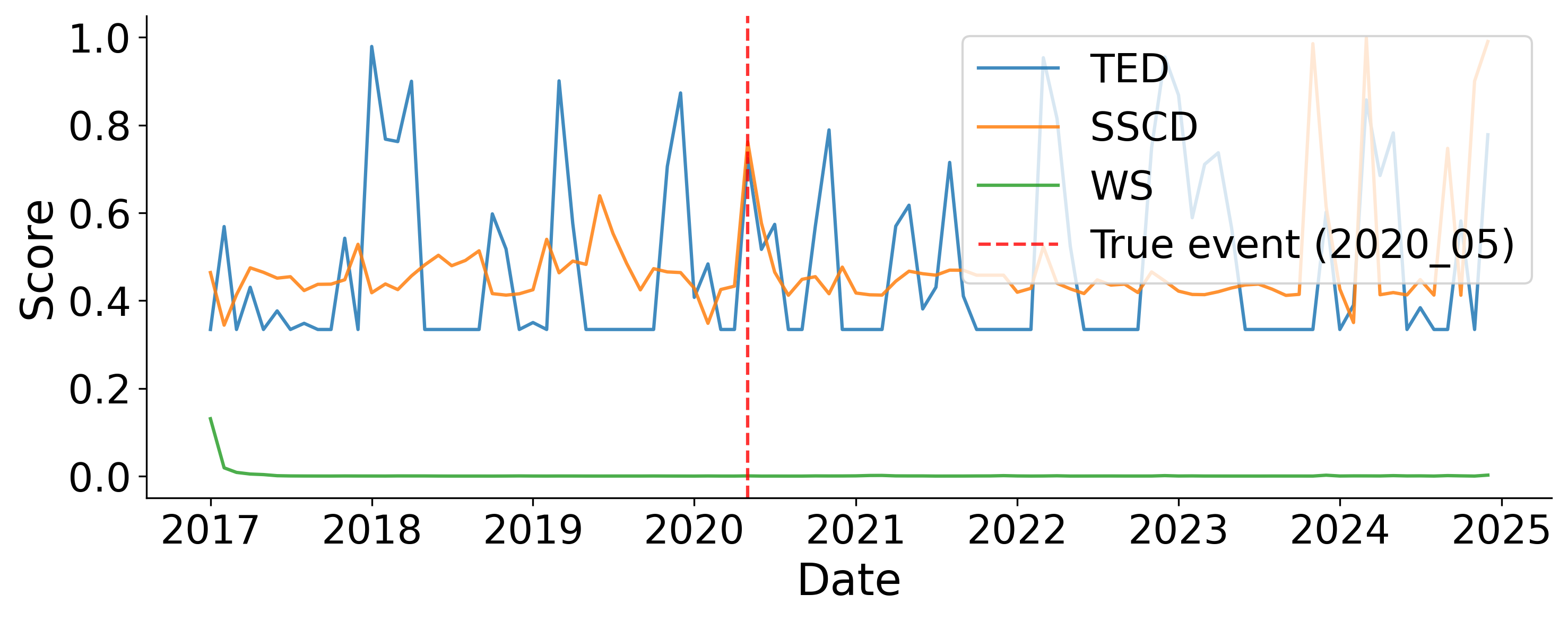}
        \caption{Site looted\_769 (event: May 2020)}
    \end{subfigure}

    \begin{subfigure}[b]{0.45\textwidth}
        \centering
        \includegraphics[width=\textwidth]{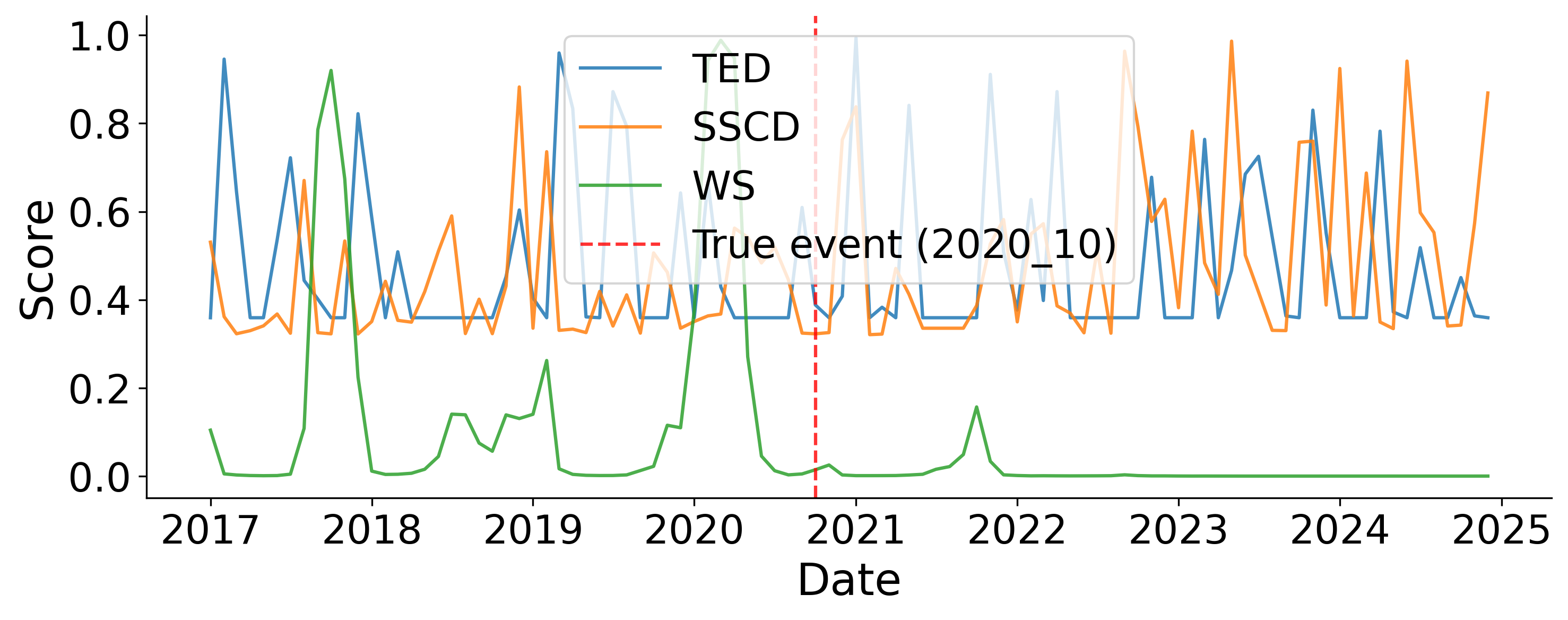}
        \caption{Site looted\_789 (event: October 2020)}
    \end{subfigure}
    \hfill
    \begin{subfigure}[b]{0.45\textwidth}
        \centering
        \includegraphics[width=\textwidth]{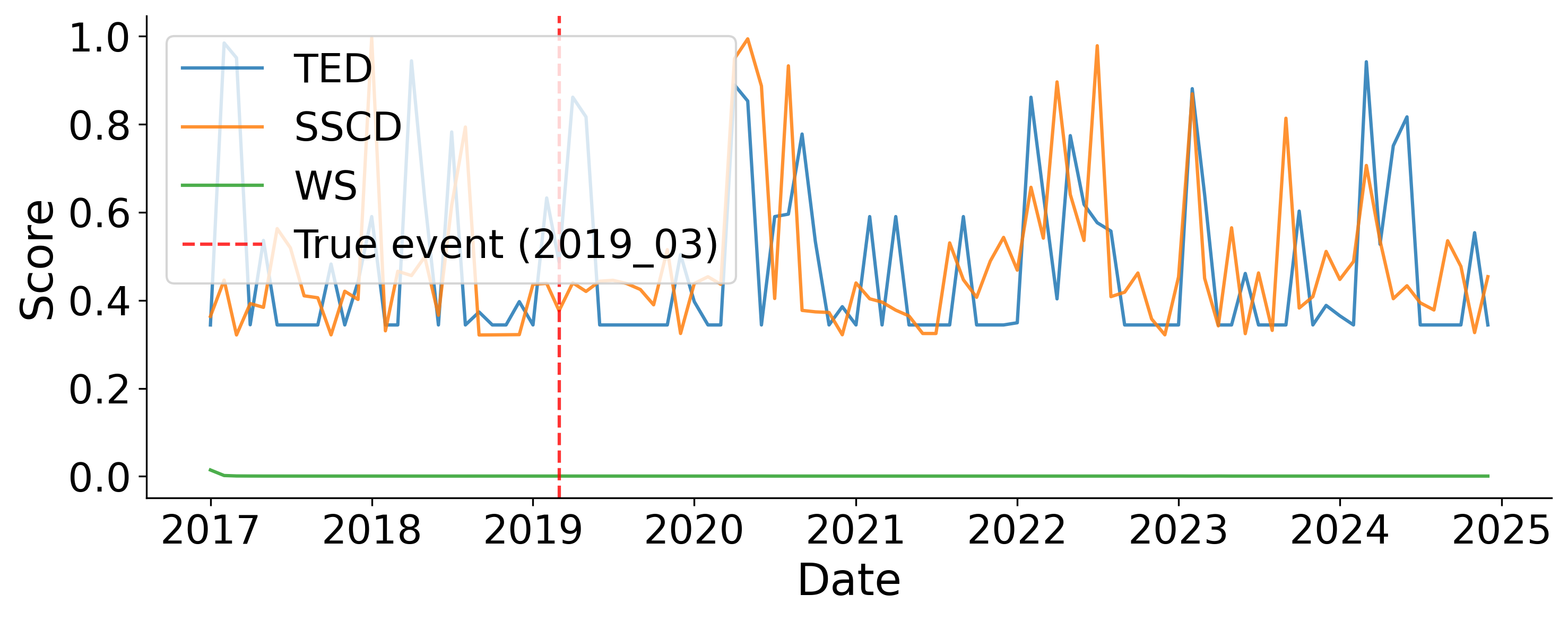}
        \caption{Site looted\_740 (event: March 2019)}
    \end{subfigure}
    \caption{Monthly GeoRSCLIP scores under three scoring regimes (TED, SSCD, WS) for four looted sites with known disturbance dates. The red dashed line marks the recorded looting month. TED and SSCD exhibit broad temporal sensitivity with peaks near the event, while WS produces sharp spikes only when the event falls within the temporal range of training labels.}
    \label{suppfig:site_examples}
\end{figure*}

Fig.~\ref{suppfig:site_examples} shows GeoRSCLIP monthly scores under all three scoring regimes for four representative looted sites with known disturbance dates.
TED and SSCD produce comparable, spiky score trajectories that respond to both genuine change events and nuisance variability throughout the timeline.
In several cases unsupervised methods generate elevated scores near the recorded looting month, illustrating their capacity for temporal localization even without labeled data.
SSCD tends to yield slightly smoother peaks than TED, consistent with its learned reconstruction objective filtering out high-frequency noise.
WS scores, by contrast, are near-zero for most of the timeline but produce sharp, isolated spikes concentrated around the event period when the true month falls within the temporal range of training labels (sites~a,~c).
When the event falls outside that range (sites~b,~d), WS remains flat, underscoring its dependence on the distribution of supervised labels rather than on general anomaly detection.

\section{Feature-Space Variability Over Time}
\label{app:feature_variation}

\begin{figure}[htbp]
\centering
\includegraphics[width=\columnwidth]{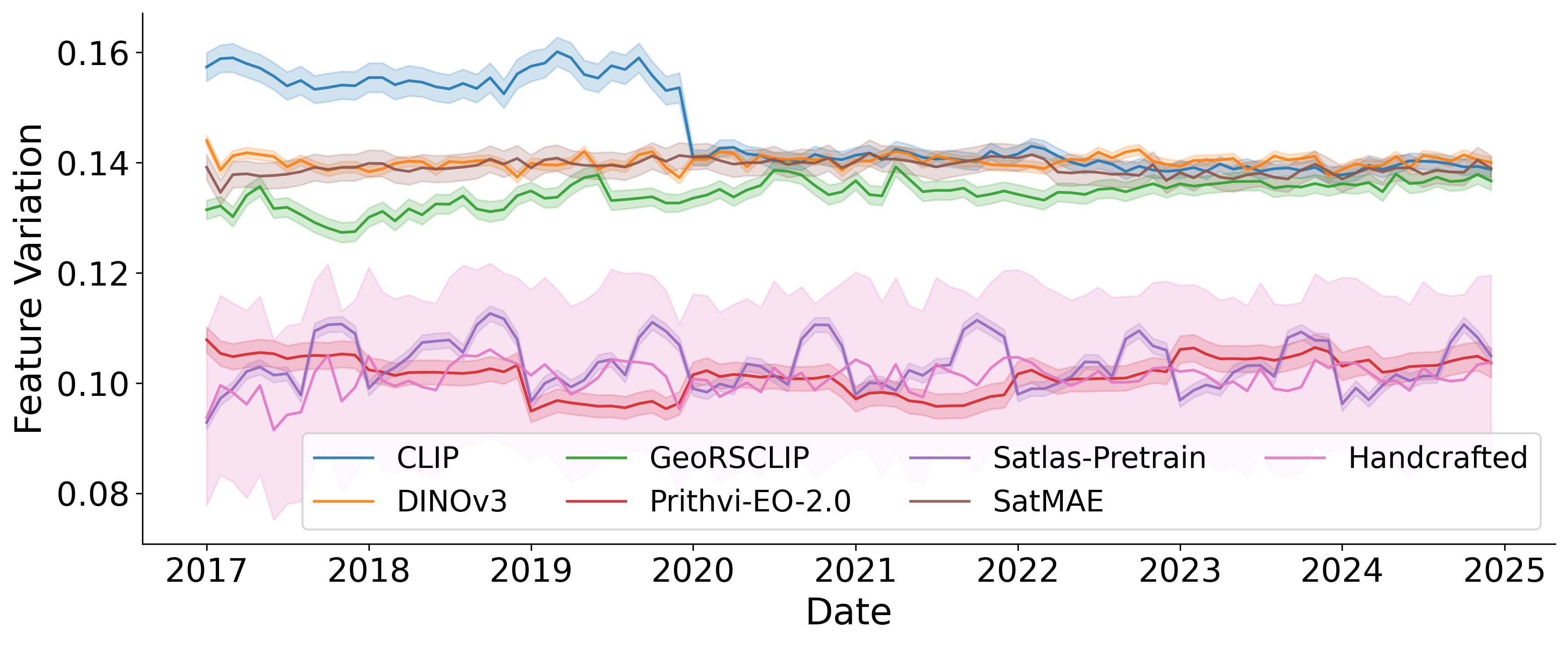}
\caption{Feature variation over time across all Afghanistan sites.
Per month, each feature dimension is min-max normalized to $[0,1]$ across sites and missing months are forward-filled; the plotted value is the mean per-feature standard deviation.
Higher values indicate greater cross-site variability in the embedding space for that month.}
\label{suppfig:feature_variation}
\end{figure}

To identify which segments of the time series are intrinsically more variable in feature space (independent of any change-label signal), we compute a feature-variation statistic for each month: (i)~min-max scale each feature dimension to $[0,1]$ across the evaluated set, (ii)~compute the standard deviation of each feature across sites, and (iii)~average these per-feature standard deviations to obtain a scalar variation score.
Fig.~\ref{suppfig:feature_variation} reveals three distinct tiers of cross-site feature variability.
CLIP exhibits the highest variation ($\sim$0.155--0.160) during 2017--2019 before dropping to $\sim$0.14, suggesting a shift in the underlying imagery distribution around that period.
DINOv3, GeoRSCLIP, and SatMAE form a middle tier with stable variation around 0.13--0.14 throughout the time series, indicating consistent but moderate sensitivity to site-level differences.
Prithvi-EO-2.0, Satlas-Pretrain, and Handcrafted cluster in a lower tier ($\sim$0.09--0.11), with Satlas-Pretrain showing a distinctive seasonal oscillation and Handcrafted displaying the most erratic month-to-month fluctuations despite its low overall level.

\end{document}